\documentclass{article}

\usepackage[preprint]{neurips_2019}

\usepackage[utf8]{inputenc} 
\usepackage[T1]{fontenc}    
\usepackage{hyperref}       
\usepackage{url}            
\usepackage{booktabs}       
\usepackage{amsfonts}       
\usepackage{nicefrac}       
\usepackage{microtype}      
\usepackage{graphicx}
\usepackage{multirow}
\usepackage{subcaption}
\usepackage{wrapfig}
\usepackage[titletoc]{appendix}
\usepackage{color}

\title{Interpretability Beyond Classification Output: Semantic Bottleneck Networks} 

\author{
  Max Losch$^1$ \quad Mario Fritz$^2$ \quad Bernt Schiele$^1$\\
  $^1$Max Planck Institute for Informatics\quad $^2$CISPA Helmholtz Center for Information Security\\
  Saarland Informatics Campus, Germany\\
  $^1$\textit{firstname.lastname@mpi-inf.mpg.de}\\$^2$\textit{firstname.lastname@cispa.saarland}}

\begin{document}

\maketitle

\begin{abstract}
Today's deep learning systems deliver high performance based on end-to-end training. While they deliver strong performance, these systems are hard to interpret. To address this issue, we propose Semantic Bottleneck Networks (SBN): deep networks with semantically interpretable intermediate layers that all downstream results are based on. As a consequence, the analysis on what the final prediction is based on is transparent to the engineer and failure cases and modes can be analyzed and avoided by high-level reasoning. We present a case study on street scene segmentation to demonstrate the feasibility and power of SBN. In particular, we start from a well performing classic deep network which we adapt to house a SB-Layer containing task related semantic concepts (such as object-parts and materials). Importantly, we can recover state of the art performance despite a drastic dimensionality reduction from 1000s (non-semantic feature) to 10s (semantic concept) channels. Additionally we show how the activations of the SB-Layer can be used for both the interpretation of failure cases of the network as well as for confidence prediction of the resulting output. For the first time, e.g., we show interpretable segmentation results for most predictions at  over 99\% accuracy.
\end{abstract}

\section{Introduction}
\vspace{-5pt}
\begin{figure}[h]
\begin{center}
  \includegraphics[width=0.9\linewidth]{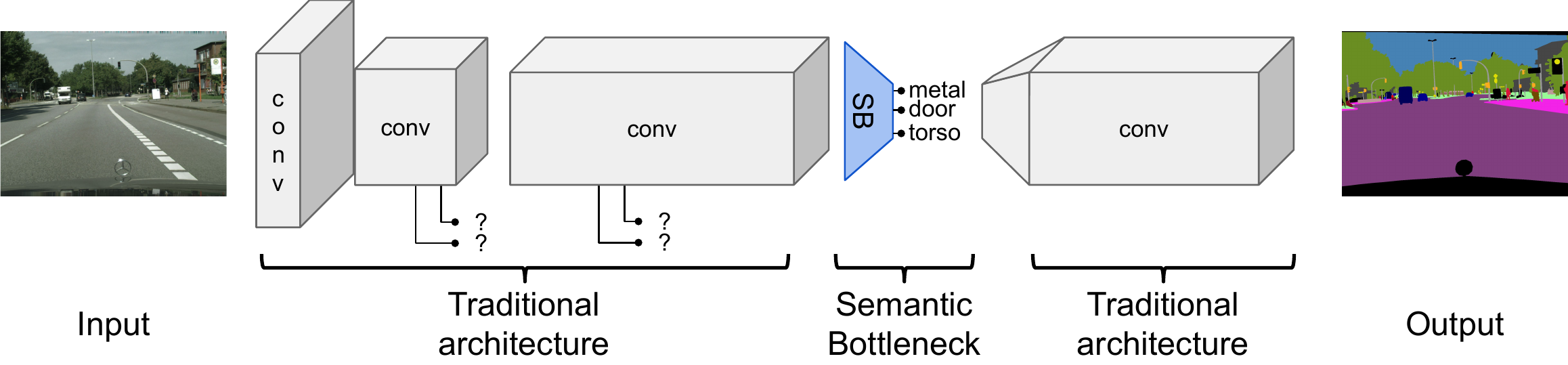}
\end{center}
\vspace{-10pt}
  \caption{Semantic Bottleneck Network (SBN). While the semantics of features in traditional architectures are unknown,
  our proposed SBN lends itself to interpretability beyond the output.}
\label{fig:intro}
\vspace{-10pt}
\end{figure}

While end-to-end training is key to the success of deep learning in terms of performance  -- it is also the main obstacle to obtain interpretable results and inspectable systems. 
A key problem is that all intermediate representations are learned data-driven and thus are not directly inspectable by humans. 
We reconcile the competing goals of end-to-end training and inspectability by proposing Deep Learning architectures with Semantic Bottlenecks (SBs) as intermediate layers. 

The construction of intermediate representations that are semantically meaningful is achieved by utilizing additional supervision  that direct the formation of representations to semantic meanings.
The approach we propose in this paper is simple yet effective: 
Assuming we have a deep neural network that performs well on our target domain, we retrospectively build our SB by learning a function that maps the original feature space onto a semantic space of our choosing.
To subsequently ensure that all downstream results are based on these semantics only, a SB-Layer is inserted into the original model
(see e.g. figure~\ref{fig:intro}).

In computer vision scenarios that we focus on in this paper, it is sensible to choose semantic representations that we expect to be important such as objects, boundaries, textures, materials, shapes, flow or depth.
From related work on analyzing semantic content in state-of-the-art architectures, we know that representations can already -- at least partially --  be associated wit object, texture or color detectors~\cite{zeiler2014visualizing,yosinski2015understanding,Zhou2015ObjectDE,bau2017network}.
We consequently expect that a mapping to relevant semantic content is feasible without impairing the performance of the original model.

Our \textbf{contribution} is four fold.
Firstly, we propose the introduction of Semantic Bottlenecks (SBs) into state-of-the-art architectures and show that this does not impair their performance, even for low-dimensional SBs. 
In fact, based on our knowledge, we are first to show an inherently interpretable model being competitive with state of the art performing classification models.
Secondly, the activations in the SB-Layer enable interpreting error cases. We show that misclassifications can often be ascribed to cases of missing or conflicting evidence and 
thirdly show that directed manipulation of evidence in the SB opens a new door for testing hypotheses like: ``does adding this semantic evidence flip the classification result to the correct label?''.
Lastly, we present a simple way of estimating the confidence for a prediction of the SBN and show that this allows to achieve a prediction accuracy of over $99\%$ for over $75\%$ of all samples.

\section{Related Work}

As argued in prior work~\cite{lipton2018mythos}, interpretability can be largely approached in two ways.
The first being post-hoc interpretation, for which we take an already trained and well performing model and dissect its decisions a-posteriori to identify important input features via attribution~\cite{bach2015pixel, selvaraju2017grad, kindermans2017reliability, zintgraf2017visualizing, shrikumar2017learning, sundararajan2017axiomatic} or attempt to assign meaning to single or groups of features in order to understand the inner workings~\cite{simonyan2013deep, zeiler2014visualizing, yosinski2015understanding, bau2017network, kim2018interpretability}.

The second approach is to start off with models that are inherently interpretable.
Inherent interpretability has just recently gained more attention with more work focusing on making the predictions based on human interpretable concepts or prototypes.
Similar to our proposed approach, \cite{al2017contextual, melis2018towards} and \cite{li2018deep} embed an interpretable layer into the network, yet are different in the way that they treat the concept representations as free parameters and train them jointly with the network parameters.
The pitfall with all three methods is, that while an automatic process is desirable, it is not guaranteed that the learned representations are in fact human interpretable. 
We circumvent that problem and simplify it drastically by forcing the network to express its information via an interpretable basis that we fix apriori, making sure that each representation has a clear meaning to begin with.

The idea of basing classification on a predetermined interpretable basis is related to  Object Bank introduced in 2010~\cite{li2010object}.
Here, the authors compose a fixed set of independent object detectors based on simple feature descriptors like SIFT~\cite{lowe2004distinctive} and SURF~\cite{bay2006surf} and perform the final classification task on the collection of all detector outputs.
In comparison, to retain state of the art performance for street scene segmentation, we train our SB on pretrained features from a network that has been shown to reach competitive performance and inject the detectors back into the network as SB.

\section{Semantic Bottleneck Networks for Interpretability and Inspection}
\label{sec:sbns}

\subsection{Construction of Semantic Bottleneck Networks}
\label{sec:sbn:construction}

\begin{wrapfigure}{R}{0.5\textwidth}
\vspace{-30pt}
    \begin{center}
       \includegraphics[width=0.5\textwidth]{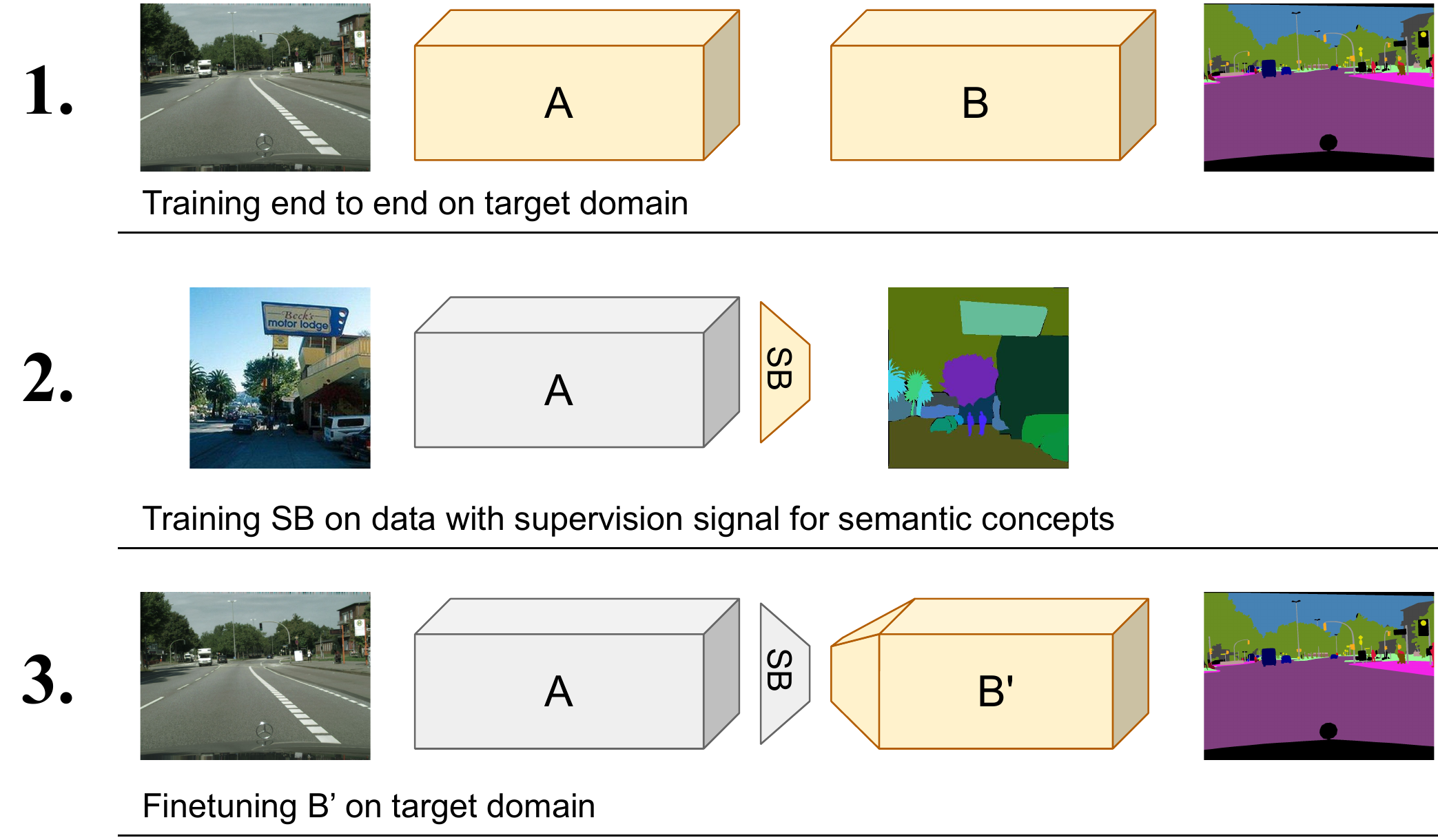}
    \end{center}
    \vspace{-5pt}
   \caption{Construction of SBNs. 1. Start off with a well performing model on the target task. 2. Train a function (SB) that maps intermediate representations to semantic concepts. 3. Insert the SB back into the original model and finetune all downstream layers.}
   \label{fig:method}
   \vspace{-18pt}
\end{wrapfigure}

To eventually end up with a SBN that is competitive with state-of-the-art models and is interpretable, we strive to find a set of semantic concepts that is rich enough to allow for good performance, yet is as small as possible to simplify inspection.
Finding the optimal set of concepts from might be impractical in general, yet we show later in our case study that a manual task-specific selection can be sufficient to satisfy both desiderata.

Given a set of selected semantic concepts, there are different options to construct the SBN.
One is training from scratch with an additional training loss on the SB-Layer to enforce high correlation between outputs and semantic meaning.
This method requires particular care during training to balance all losses involved.
An alternative is to start from a well-performing network and inject a SB in a two step procedure.
First, as shown by point $2$ in figure~\ref{fig:method}, we train a function that maps the representations of an intermediate layer to our desired semantic space keeping the host model (\textit{A}) fixed.
Then, as indicated by point $3$ of the same figure, we insert the SB and finetune the downstream layers (\textit{B'}) to accommodate for the changed feature space, holding \textit{A} and \textit{SB} fixed.
To show the power of SBNs, in this paper, we focus on the latter method.

\subsection{Inspection of error modes}
\label{sec:sbn:erroranalysis}

The power of SBNs lies in the ability to inspect the evidence for the chosen semantic concepts to investigate errors.
Such errors could involve the absence of evidence for a particular class indicating that the upstream information was lost or never extracted.
In contrast, it is also possible to detect that there is competing evidence from orthogonal concept categories, or even superseded by them, indicating that the downstream classifier made the wrong decision in the end or the upstream layers integrated the low level evidence incorrectly.
We will come back to such errors in the experiment section~\ref{sec:exp:erroranalysis} and will discuss them in more depth.

Besides investigating single samples, the semantic concepts allow to study the errors made by the SBN more globally.
In particular, we are interested in finding patterns of errors that can be used to further increase the understanding on how the SBN processes information.
Importantly, finding such patterns could allow the construction of simple high level fail-safe systems in order to avoid misclassifications during runtime.

\begin{figure}[h]
    \vspace{-10pt}
    \begin{center}
       \includegraphics[width=0.7\linewidth]{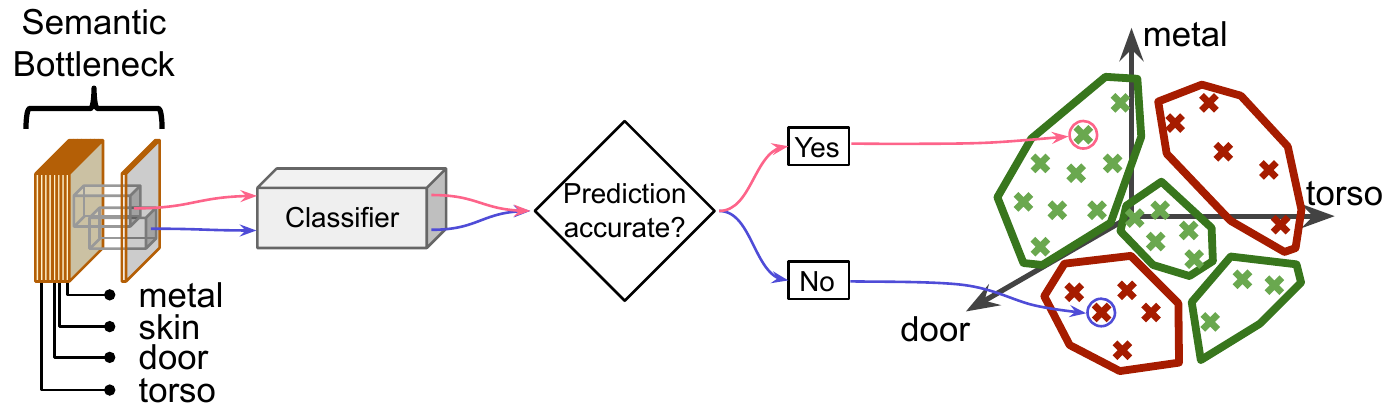}
    \end{center}
    \vspace{-10pt}
    \caption{Populating the SB space to find modes of errors. The gray boxes enclosed in the SB indicate the receptive field of the classifier.}
    \label{fig:sb_cluster}
    \vspace{-15pt}
\end{figure}

In this paper, we approach the global error mode inspection via utilizing agglomerative clustering~\cite{ward1963}, enabling the investigation of errors on the cluster level and sample level as well contrasting errors with the closest true positive cluster (see figure~\ref{fig:sb_cluster} for a schematic of the method).
The latter being a safe substitute for counterfactuals~\cite{wachter2017counterfactual}, which could lead to unwanted adversarial examples~\cite{szegedy2013intriguing, goodfellow2014explaining, kurakin2016adversarial, madry2017towards}.

\subsection{Estimating prediction confidence}
\label{sec:sbn:confidence}

Given the high performance that today's deep learning systems deliver, it is of increasing interest to deploy such systems in the real world.
Besides inspection of error modes, it is highly desirable to acquire an estimate of confidence for every prediction the system makes.
Thus, we investigate the suitability of the semantic representations in the SB for uncertainty prediction by training shallow networks directly on the SB conditioned on whether the class is present at a given pixel or not.
Our approach is therefore an approximation of the complex decision boundary of a deep net, giving us insights into how descriptive the representations in the semantic bottleneck are with respect to each target class and how assertive the model can be.
Depending on the concepts in our SB, we expect that some classes are easier to be certain about than others, which will be represented by higher confidence scores.
This in turn enables us to acquire meta information on which classes are difficult to predict based on the information given, again enabling the construction of fail-safe methods, and more importantly to enable further iterations on the choice of concepts in the SB.

\subsection{Case study: Street scene segmentation}

To build support for our claims, we subsequently transfer the introduced methods into practice by applying them to our case study: semantic segmentation.
While semantic scene segmentation has attained  high  performance, significant failure modes remain and basically each result image   contains errors easily spotted by a human observer.
For our case study we use the Cityscapes dataset~\cite{cordts2016cityscapes} which consists of $19$ different classes, $5,000$ images  with fine grained labeling and $20,000$ with coarse labeling as well as $500$ validation images.
Many pretrained architectures are available. We use PSPNet~\cite{zhao2017pyramid} based on  ResNet-101~\cite{he2016deep},   
due to its strong performance and its sequential order of processing.
The latter simplifying the introduction of SBs.

\begin{wrapfigure}[21]{R}{0.22\linewidth}
\vspace{-15pt}
\centering
\begin{subfigure}[t]{\linewidth}
    \begin{center}
      \includegraphics[width=\linewidth]{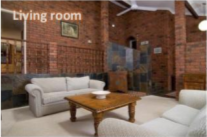}
    \end{center}
    \vspace{-10pt}
\end{subfigure}

\begin{subfigure}[t]{\linewidth}
    \begin{center}
      \includegraphics[width=\linewidth]{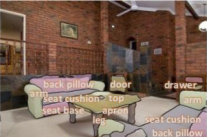}
    \end{center}
    \vspace{-10pt}
\end{subfigure}

\begin{subfigure}[t]{\linewidth}
    \begin{center}
      \includegraphics[width=\linewidth]{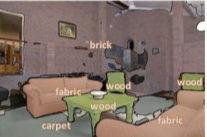}
    \end{center}
    \vspace{-10pt}
\end{subfigure}
\vspace{-15pt}
\caption{Sample from Broden+ dataset with annotations for parts (2nd row) and materials (3rd).}
\label{fig:broden_sample}
\end{wrapfigure}

As discussed, we want to learn relevant semantic representations in our SB with additional supervision.
\textit{Broden+}~\cite{xiao2018unified} is a recent collection of datasets which serves as a starting point of our case study as it contains annotations for a broad range of relevant semantic concepts.
It offers thousands of images for objects, parts, materials and textures for which the first three types come with pixel level annotations (see figure~\ref{fig:broden_sample} for an example) of which many are relevant for the $19$ classes in the Cityscapes dataset.

Based on the $377$ part and material concepts available ($351$ parts sourced from ADE~\cite{zhou2017scene} and Pascal-Part~\cite{chen2014detect} and $26$ materials sourced from OpenSurfaces~\cite{bell2013opensurfaces}), we compile multiple subsets containing at least $6$ concepts in order to explore the relationship between number of concepts and segmentation performance on Cityscapes.
In particular, many of the $377$ concepts are not relevant for the task of street scene segmentation, which we anticipate to be not important for reaching good performance and interpretability.
Consequently, we split the concept set into $70$ task-relevant and $307$ task-irrelevant concepts in order to test the effect of task-relatedness on the model's performance (see table~\ref{tab:selection_of_relevant} for the relevant).
A detailed list of used concepts is presented in table~\ref{tab:concepts} in the appendix.
We abstain from using any object concepts that are on the same abstraction level as the Cityscapes classes (e.g. wheel in \textit{Broden+} is not defined as standalone object but as part of the object car), which would give us little added benefit for inspection, but also would just shift the segmentation task to an earlier layer.
Instead we are particularly interested in selecting concepts that can be thought of as subordinate attributes to the target classes.

\begin{table}
\begin{tiny}
    \centering
    \begin{tabular}{l|c|c|c|c|c|c|c|c|c|c|c|c}
        \textit{Broden+} object & Sky & Building & Person & Road & Car & Lamp & Bike & Van & Truck & Motorbike & Train & Bus \\\hline
        \# subordinate parts & 1 & 5 & 14 & 1 & 9 & 3 & 4 & 6 & 2 & 3 & 5 & 6\\\hline\hline
        \textit{Broden+} material & \multicolumn{12}{c}{Brick,  Fabric,  Foliage,  Glass,  Metal,  Plastic,  Rubber, Skin,  Stone,  Tile,  Wood}
    \end{tabular}
\end{tiny}
    
    \caption{Relevant concepts from \textit{Broden+} for the Cityscapes domain. Material concepts in bottom row and parts are grouped by their respective parent object (top 2 rows).
    }
    \label{tab:selection_of_relevant}
    \vspace{-20pt}
\end{table}

\subsection{Implementation details}
\label{sec:sbn:implementation}

Following the method of starting off with a pretrained and well performing network, we first train the function that maps the original representations to our semantic concepts by training concept detectors.
Part and and material detectors are trained separately with one cross entropy loss each, as a single image pixel can have both part and material annotations.
We choose simple linear classifiers for this task, which we train with stochastic gradient descent, a batchsize of $16$ which we split into $2$ sub-batches per gpu and perform gradient accumulation to perform the final update step.
The learning rate is set to $0.002$, the weight decay to $5e-4$ and the parameters are trained for $5,000$ iterations with ``poly'' learning rate policy~\cite{chen2018deeplab, zhao2017pyramid, xiao2018unified}.

Integrating the SB with the host network now involves adjusting the dimensionality of the first layer of part \textit{B} (as in step $3$ of figure~\ref{fig:method}) to fit the dimensionality of the SB.
Afterwards, all downstream layers are finetuned following the training details from the original PSPNet paper~\cite{zhao2017pyramid} with some small adjustments due to restrictions of our hardware resources.
Instead of using an input cropsize of $713$x$713$ we have to limit the cropsize to $651$ for our training. 
The hyperparameters are otherwise the same as for the training of the SB except for the number of iterations, which we increase to $10,000$.
For all our finetunings, we do not employ the deep supervision as is originally described.

\section{Experiments}

We now discuss experimental results consisting of constructing various SBNs with different numbers of related and unrelated concepts (section~\ref{sec:exp:construction}), the error analysis (section~\ref{sec:exp:erroranalysis}) and evaluation of confidence prediction (section~\ref{sec:exp:confidence}).
Also, we discuss (section~\ref{sec:exp:manipulation}) an additional benefit of having access to semantic meaning in intermediate layers: manipulation of evidence.

\subsection{Choice of semantic concepts for Cityscapes}
\label{sec:exp:construction}

\begin{wraptable}{R}{0.5\textwidth}
\vspace{-30pt}
\centering
\begin{footnotesize}
\begin{tabular}{l|c|c|c}
&\#concepts & &\\
configuration &  (materials, parts) & mIoU & pAcc\\\hline\hline
{\small PSPNet} &N/A& 78.5 & 95.2\\
{\small PSPNet$\ddagger$} &N/A& 76.2 & 95.6\\\hline\hline
{\small SB@block1} & \multirow{2}{*}{70 (11, 59)} & \multirow{2}{*}{74.4} & \multirow{2}{*}{95.5}\\
{\small(128 input feat.)} & & &\\\hline

{\small SB@block2} & \multirow{2}{*}{70 (11, 59)} & \multirow{2}{*}{74.7} & \multirow{2}{*}{95.5}\\
{\small(256 input feat.)} & & &\\\hline

{\small SB@block3} & \multirow{2}{*}{70 (11, 59)} & \multirow{2}{*}{73.1} & \multirow{2}{*}{95.1}\\
{\small(512 input feat.)} & & &\\\hline

{\small SB@block4} & \multirow{2}{*}{70 (11, 59)} & \multirow{2}{*}{76.2} & \multirow{2}{*}{95.5}\\
{\small(1024 input feat.)} & & &\\\hline

{\small SB@pyramid} & \multirow{2}{*}{70 (11, 59)} & \multirow{2}{*}{72.8} & \multirow{2}{*}{94.7}\\
{\small(4096 input feat.)} & & &\\\hline

{\small SB@penultimate} & \multirow{2}{*}{70 (11, 59)} & \multirow{2}{*}{65.1} & \multirow{2}{*}{94.3}\\
{\small(512 input feat.)} & & &\\
\end{tabular}
\end{footnotesize}
\vspace{-5pt}
\caption{Segmentation results on Cityscapes validation set for different placements of the SB. PSPNet$\ddagger$ is the vanilla architecture trained for $10,000$ iterations longer.}
\vspace{-15pt}
\label{tab:sb_placement_results}
\end{wraptable}

We evaluate our finetuned models with single scale and without mirroring on the full Cityscapes validation set and present our experiments in three parts.
First of all, we construct five SBNs with $70$ task-relevant concepts at different layers in the PSPNet architecture and compare mIoU and pixel accuracy against the vanilla architecture (see table~\ref{tab:sb_placement_results}).
Given the baseline pixel accuracy of $95.2\%$, the worst performing SBN is at the second to last layer (penultimate) with $94.3\%$ and the best performing resnet block $1$ and $4$ with a slight improvement to $95.5\%$.
Moreover, the baseline mIoU performance is $78.5\%$ and for our SBNs ranges from $65.1\%$ at penultimate to $76.2\%$ at block $4$.
We observe a decrease in mIoU, but increase in pixel accuracy.
\begin{wraptable}{R}{0.35\textwidth}
\vspace{-20pt}
\centering
\begin{footnotesize}
\begin{tabular}{c|c|c}
\#concepts & &\\
(materials, parts) & mIoU & pAcc\\\hline\hline
377 (26, 351) & 76.4 & 95.7\\\hline
162 (18, 144) & 76.0 & 95.7\\\hline
70 (11, 59) & 72.8 & 94.3\\\hline
36 (6, 30) & 65.1 & 93.3\\\hline
6 (1, 5) & 26.0 & 82.1\\
\end{tabular}
\end{footnotesize}
\vspace{-8pt}
\caption{Segmentation results on Cityscapes validation set for different number of semantic concepts @pyramid}
\label{tab:sb_size_results}
\vspace{-15pt}
\end{wraptable}
This can be explained by the cross-entropy objective maximizing the pixel accuracy and not the mIoU metric.
To confirm, we continue finetuning the pretrained PSPNet without any SB (indicated in the table by $\ddagger$) and observe a decrease in mIoU to $76.2\%$ from $78.5\%$ and an increase in pixel accuracy to $95.6\%$ from $95.2\%$, therefore explaining our obtained results.

Concluding the first experiment, we see that some placements of the SB result in better performing models than others. 
While only block $1$ and $4$ are able to recover the full pixel accuracy of PSPNet, we observe that most come close to full recovery.
In particular, note that the dimensionality reduction for SB@pyramid is from $4096$ channels down to $70$, with only a surprisingly small performance loss.

We continue the investigation on the impact of the SB size in the following experiment.
We focus our attention here on the SB@pyramid configuration and train it with $5$ sets of concepts with decreasing size, starting with the complete set of $377$ concepts, down to $6$ concepts.
All sets selected by hand with increasing importance for the Cityscapes segmentation task. 
While the full concept set includes concepts like \textit{furniture} or \textit{domestic appliances}, they are removed in the subsequent subset with $162$ concepts.
The subset of size $70$ is reduced by concepts related to \textit{animals} and the further sets are reduced by a reduction of parts like \textit{eyebrow} or \textit{license plate} that seem very specific.
Shifting our attention to the performance results in table~\ref{tab:sb_size_results}, we see for $377$ concepts that both the mIoU and pixel accuracy are slightly improving over the finetuned baseline in the previous table~\ref{tab:sb_placement_results} with $76.4\%$ and $95.7\%$ respectively. 
Some continuous yet slow degradation of performance takes place when reducing the number of concepts to $36$ and only when reduced to $6$ concepts we see a drastic drop in performance with $26\%$ mIoU and $82.1\%$ pixel accuracy.
We conclude that the performance is fairly robust with respect to the number of concepts and, encouragingly, does approach the baseline performance quickly.

\begin{wrapfigure}[10]{R}{0.3\textwidth}
    \vspace{-3em}
    \centering
    \includegraphics[width=\linewidth]{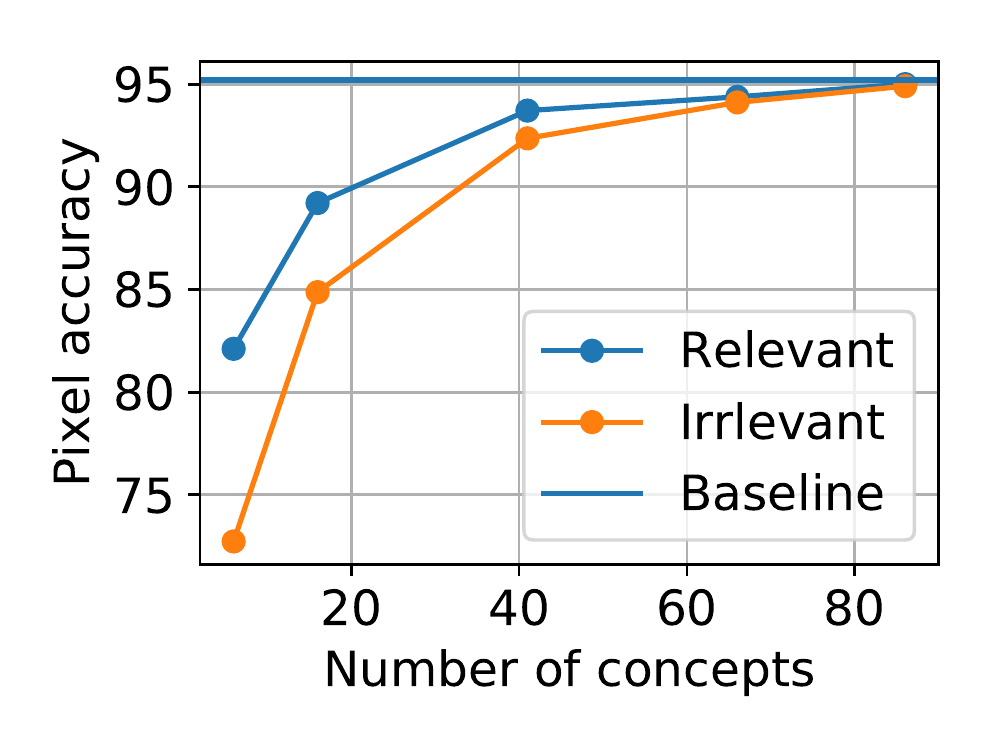}
    \vspace{-15pt}
    \caption{Task relevant concepts outperform irrelevant ones.}
    \label{fig:related_concepts}
    \vspace{-15pt}
\end{wrapfigure}
Finally, we are interested whether manual selection of task relevant concepts outperforms a random selection of irrelevant concepts.
We report the pixel accuracy results for $5$ different set sizes from $6$ to $86$ concepts (chosen in a similar way as before) in figure~\ref{fig:related_concepts} and find indeed that the selection of relevant concepts results in improved numbers, especially when the number of concepts is small (here e.g. below $60$).
Starting from $86$ concepts, the difference becomes negligible. 
We conjecture that this happens when the set of concepts is over complete.

\subsection{Semantic losslessness between SB and output}
\label{sec:exp:manipulation}

An experiment that we find necessary to conduct as sanity check, is the inspection of whether the relationship between semantic content and classes make sense, whether the feed forward pass from semantically meaningful concepts to the final network output is ``semantically lossless''.
This can be examined via the newly gained ability to manipulate the SB at will in order to observe any change in predictions.
The procedure is simple.
As an example, set any positive evidence for all \textit{building} related concepts to $0$ and reevaluate the output.
If the network has learned a semantically meaningful relationship between SB and output, all or most pixels of class \textit{building} should eventually be misclassified.

\begin{wraptable}[13]{R}{0.5\textwidth}
\vspace{-20pt}
\centering
\begin{footnotesize}
\setlength{\tabcolsep}{2pt}
\begin{tabular}{c|l|c c c c c c}
& rm concepts & build & wall & car & truck & mbike & bike\\\hline
&none & 91.4 & 51.4 & 93.9 & 81.1 & 63.2 & 74.1 \\\hline\hline
\parbox[t]{3mm}{\multirow{4}{*}{\rotatebox[origin=c]{90}{block4}}}
& build related & \textbf{0.0} & \textbf{2.8} & 93.2 & 67.5 & 58.2 & 70.4\\
& car related & 90.4 & 33.4 & \textbf{0.0} & \textbf{5.6} & 59.6 & 71.6\\
& mbike related & 91.2 & 47.6 & 93.5 & 80.1 & \textbf{8.9} & 72.0\\
& bike related & 91.1 & 49.3 & 93.7 & 81.0 & \textbf{11.6} & \textbf{7.7}\\\hline
\hline
\parbox[t]{3mm}{\multirow{4}{*}{\rotatebox[origin=c]{90}{pyramid}}}
& build related & \textbf{25.5} & \textbf{36.8} & 92.5 & 70.1 & 58.5 & 72.6\\
& car related & 90.3 & 48.3 & \textbf{25.7} & \textbf{27.9} & 56.4 & 73.5\\
& mbike related & 90.8 & 50.3 & 93.2 & 70.9 & \textbf{22.4} & 73.0\\
& bike related & 90.9 & 49.2 & 93.1 & 71.2 & \textbf{32.8} & \textbf{39.1}\\
\end{tabular}
\end{footnotesize}
\caption{Per-class mIoU results on Cityscapes validation set after removing concepts related to a particular class. Numbers in bold mark greatest degradations.}
\label{tab:ablation_results}
\end{wraptable}

\begin{figure}
    \centering
    \begin{subfigure}[t]{0.32\linewidth}
        \begin{center}
           \includegraphics[width=\linewidth]{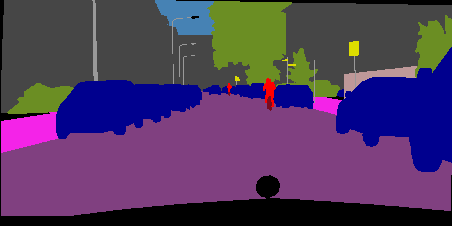}
        \end{center}
        \vspace{-10pt}
        \caption{Groundtruth segmentation}
    \end{subfigure}
    \begin{subfigure}[t]{0.32\linewidth}
        \begin{center}
           \includegraphics[width=\linewidth]{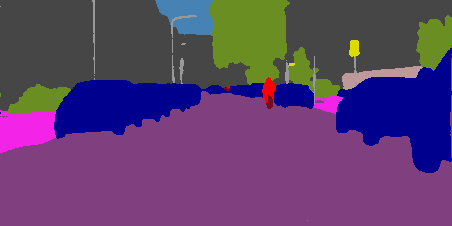}
        \end{center}
        \vspace{-10pt}
        \caption{Prediction - SB@block4}
    \end{subfigure}
    \begin{subfigure}[t]{0.32\linewidth}
        \begin{center}
           \includegraphics[width=\linewidth]{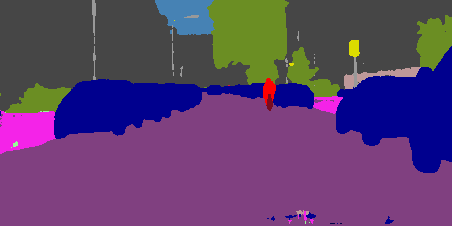}
        \end{center}
        \vspace{-10pt}
        \caption{Prediction - SB@pyramid}
        \label{fig:causal:rm_buildings:pyramid}
    \end{subfigure}

    \begin{subfigure}[t]{0.32\linewidth}
        \begin{center}
           \includegraphics[width=\linewidth]{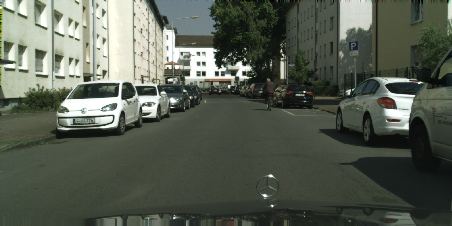}
        \end{center}
        \vspace{-10pt}
        \caption{Input image}
    \end{subfigure}
    \begin{subfigure}[t]{0.32\linewidth}
        \begin{center}
           \includegraphics[width=\linewidth]{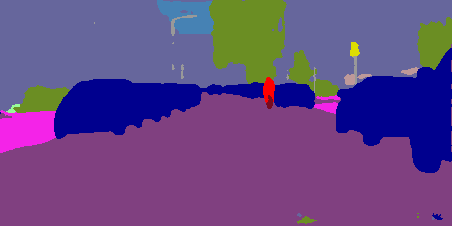}
        \end{center}
        \vspace{-10pt}
        \caption{Prediction - SB@block4 - no building evidence}
    \end{subfigure}
    \begin{subfigure}[t]{0.32\linewidth}
        \begin{center}
           \includegraphics[width=\linewidth]{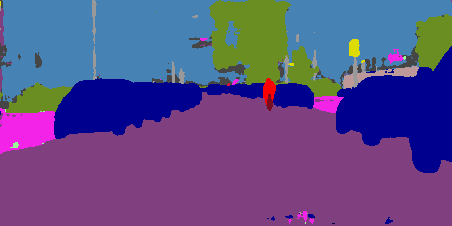}
        \end{center}
        \vspace{-10pt}
        \caption{Prediction - SB@pyramid - no building evidence}
    \end{subfigure}
    \vspace{-5pt}
    \caption{Segmentation with the SB placed at two different locations in the network results in different outputs when removing all positive evidence for building concepts. While the SB at a later layer \ref{fig:causal:rm_buildings:pyramid} fills in the building area with label \textit{sky}, the SB@block4 labels it as wall.}
    \label{fig:causal:rm_buildings}
    \vspace{-18pt}
\end{figure}

We conduct a qualitative evaluation on a single input image with the aforementioned procedure in removing concepts that relate to class \textit{building} and evaluate two different SB locations: \textit{block4} and \textit{pyramid}.
The prediction results are displayed in figure~\ref{fig:causal:rm_buildings}.
We observe that for both cases the \textit{building} pixels are misclassified and very few other pixel predictions are affected.

As one example is not representative, we repeat the experiment for four different target classes on the complete Cityscapes validation set and find agreement with the first inspection (see table~\ref{tab:ablation_results} for results). Interestingly, some concepts correlate with multiple, conceptually related classes (e.g. \textit{handle bar} and \textit{wheel} for both \textit{bike} and \textit{motorbike}) and thus the decrease is less pronounced.
Note that the network with the SB at an earlier layer is able to reach an mIoU value of $0.0$ for both \textit{building} and \textit{cars}, while a placement after the pyramid results in less strong numbers.

\subsection{Error analysis}
\label{sec:exp:erroranalysis}

Knowing that the SB networks can recover state-of-the-art performance on Cityscapes, we will now inspect the types of errors the SB@pyramid configuration with $70$ concepts makes.
We choose here for simplicity the configuration at the pyramid pooling layer as it reduces the receptive field of the classifier on the SB to $3$x$3$.
As the receptive field size is small and neighboring values are highly correlated, we average all spatial locations to acquire a sample dimensionality of $70$.
Please consult section~\ref{app:clustering} in the appendix for further details on our clustering procedure.

\begin{figure}
    \centering
    \begin{subfigure}[t]{0.59\linewidth}
        \begin{center}
           \includegraphics[width=\linewidth]{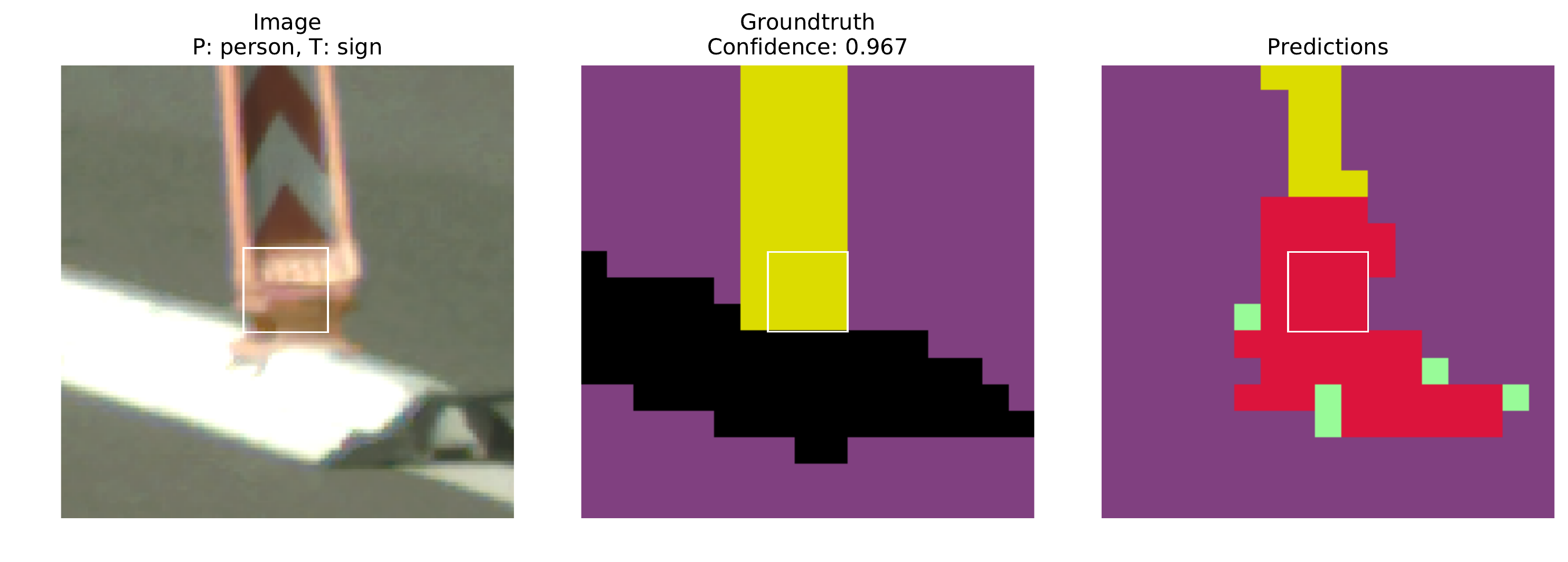}
        \end{center}
    \end{subfigure}
    \begin{subfigure}[t]{0.4\linewidth}
        \begin{center}
           \includegraphics[width=\linewidth]{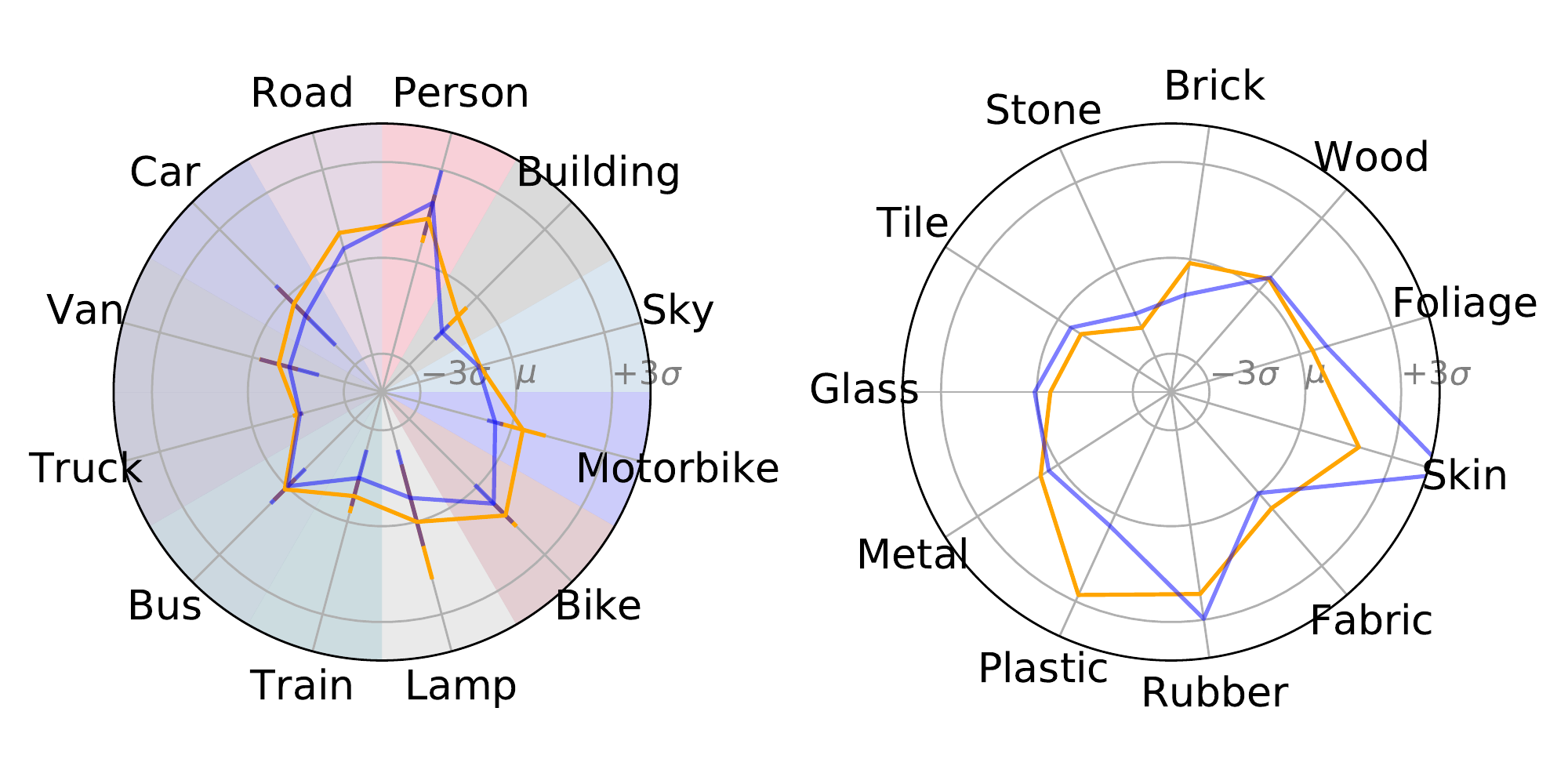}
        \end{center}
    \end{subfigure}
    
    \vspace{-15pt}
    \rule{\linewidth}{0.5pt}
    \begin{subfigure}[t]{0.59\linewidth}
        \begin{center}
           \includegraphics[width=\linewidth]{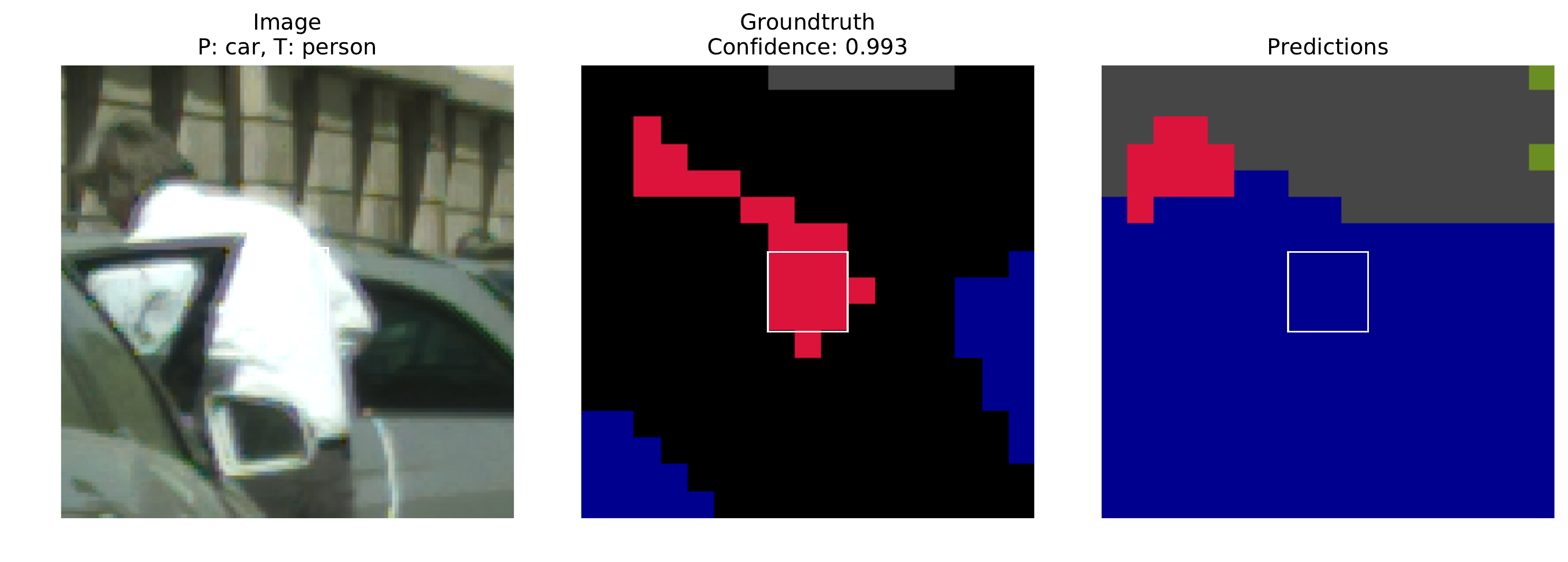}
        \end{center}
    \end{subfigure}
    \begin{subfigure}[t]{0.4\linewidth}
        \begin{center}
           \includegraphics[width=\linewidth]{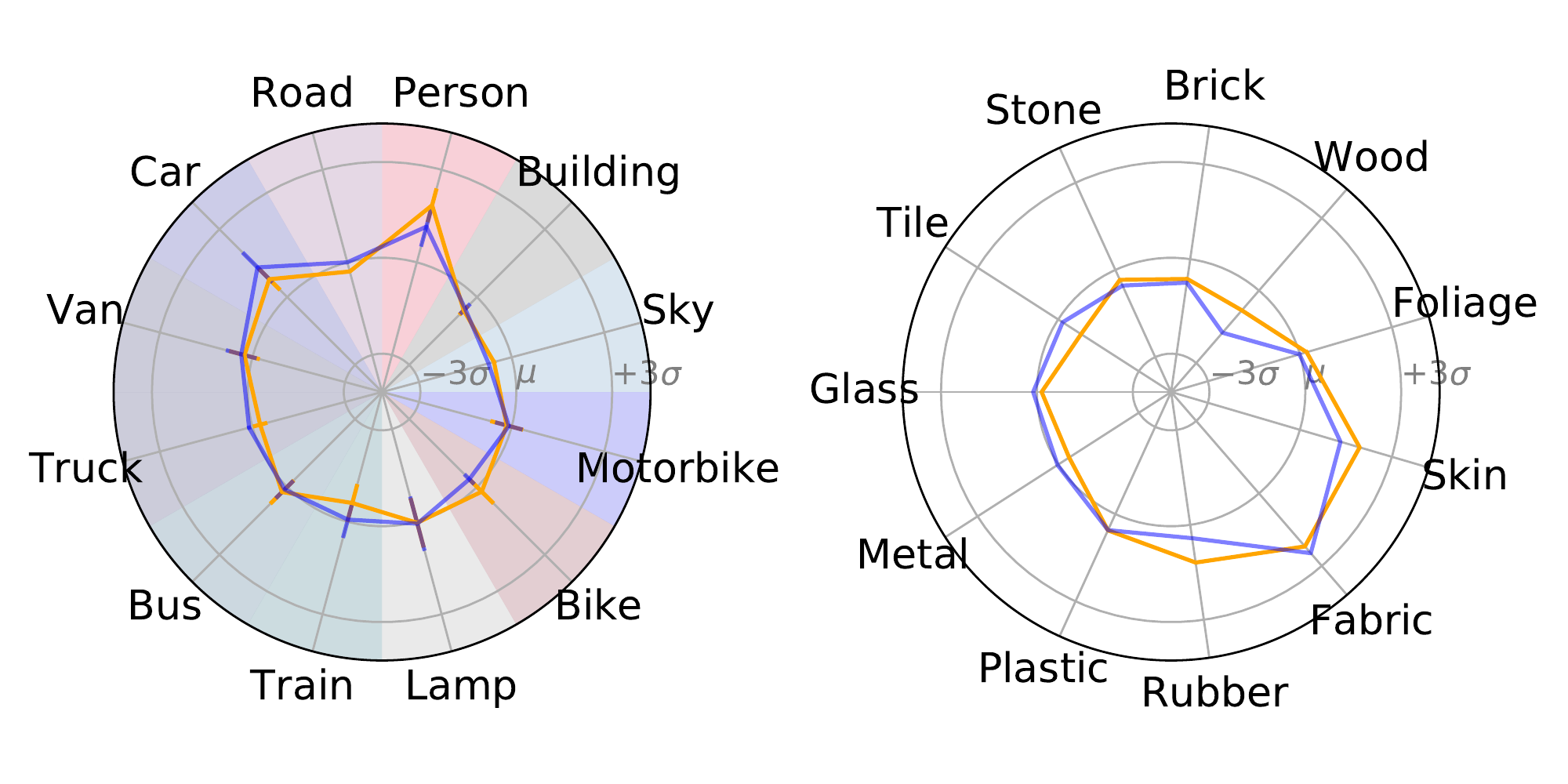}
        \end{center}
    \end{subfigure}
    
    \vspace{-15pt}
    \rule{\linewidth}{0.5pt}
    \begin{subfigure}[t]{0.59\linewidth}
        \begin{center}
           \includegraphics[width=\linewidth]{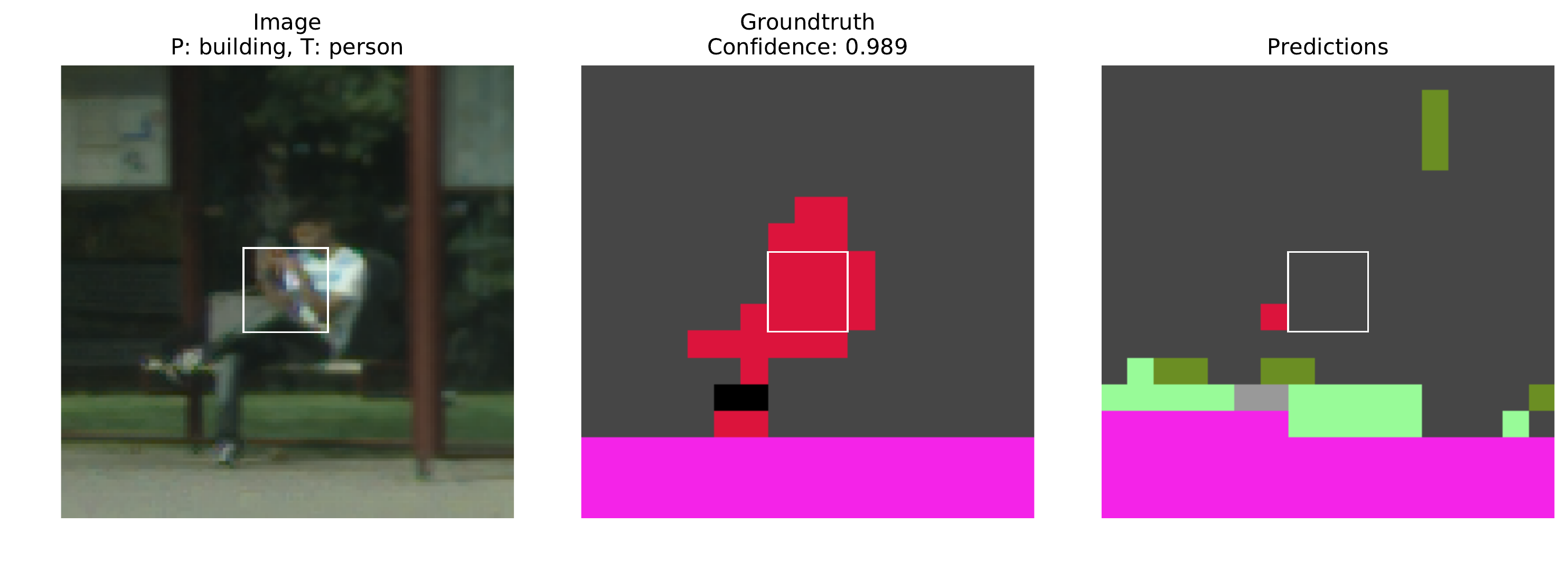}
        \end{center}
    \end{subfigure}
    \begin{subfigure}[t]{0.4\linewidth}
        \begin{center}
           \includegraphics[width=\linewidth]{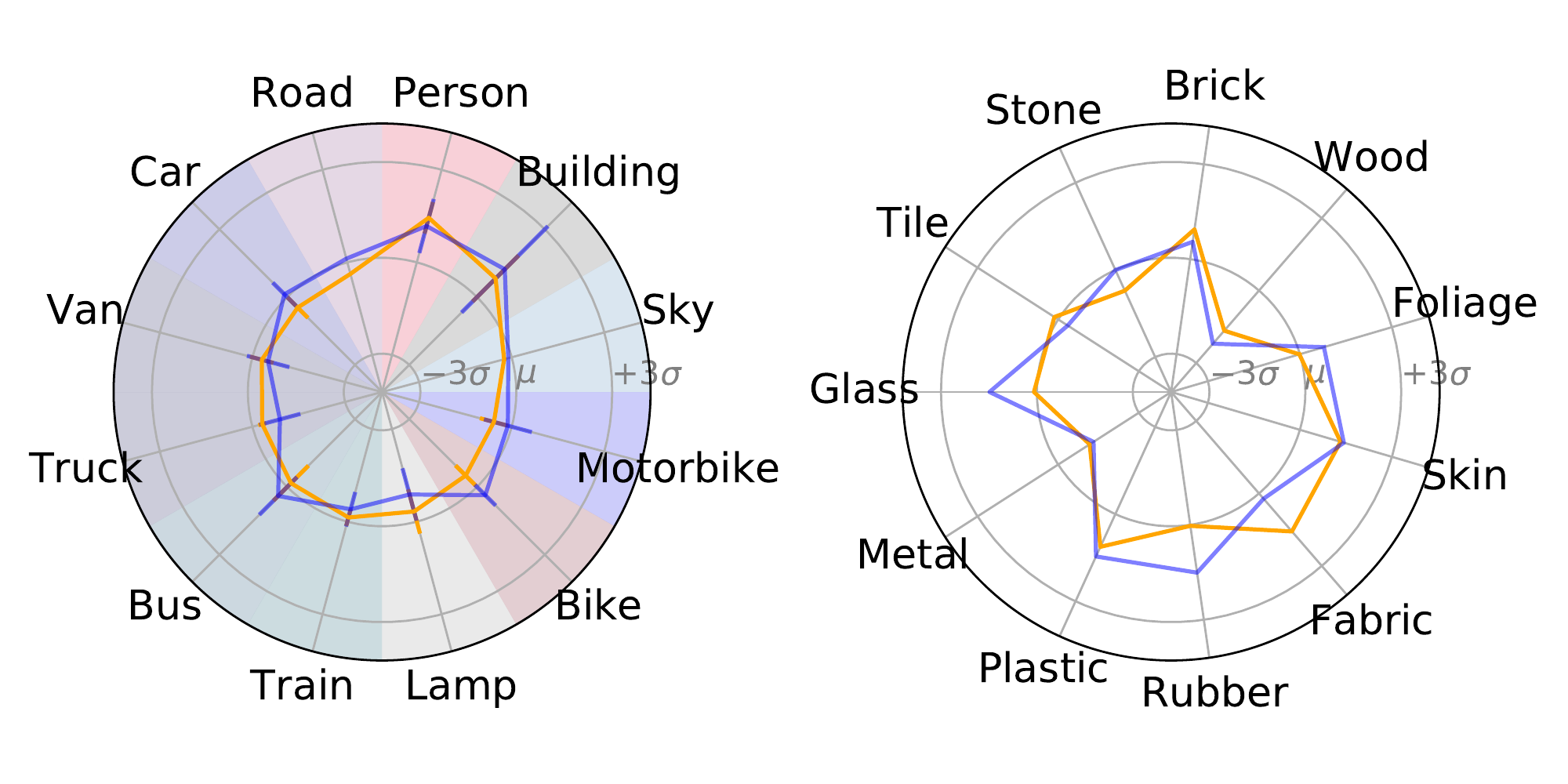}
        \end{center}
    \end{subfigure}
    
    \vspace{-15pt}
    \rule{\linewidth}{0.5pt}
    \begin{subfigure}[t]{0.59\linewidth}
        \begin{center}
           \includegraphics[width=\linewidth]{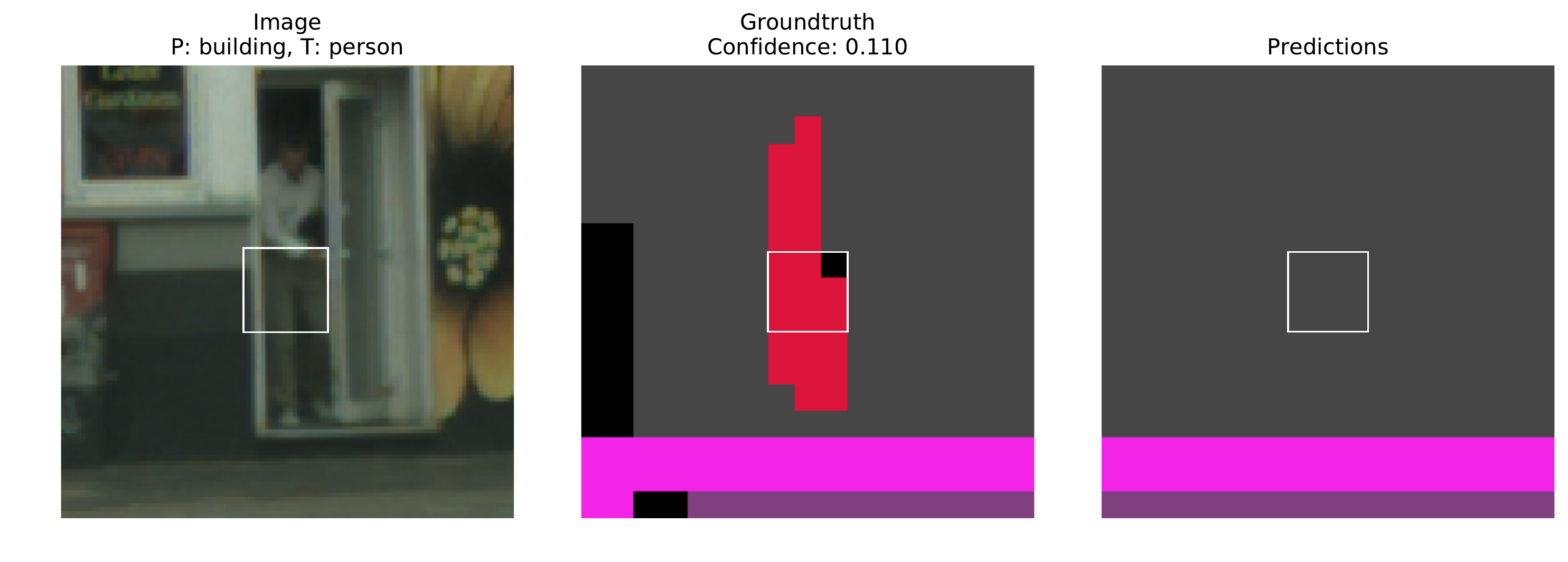}
        \end{center}
        \vspace{-10pt}
        \caption{ROI in image, groundtruth and prediction}
    \end{subfigure}
    \begin{subfigure}[t]{0.4\linewidth}
        \begin{center}
           \includegraphics[width=\linewidth]{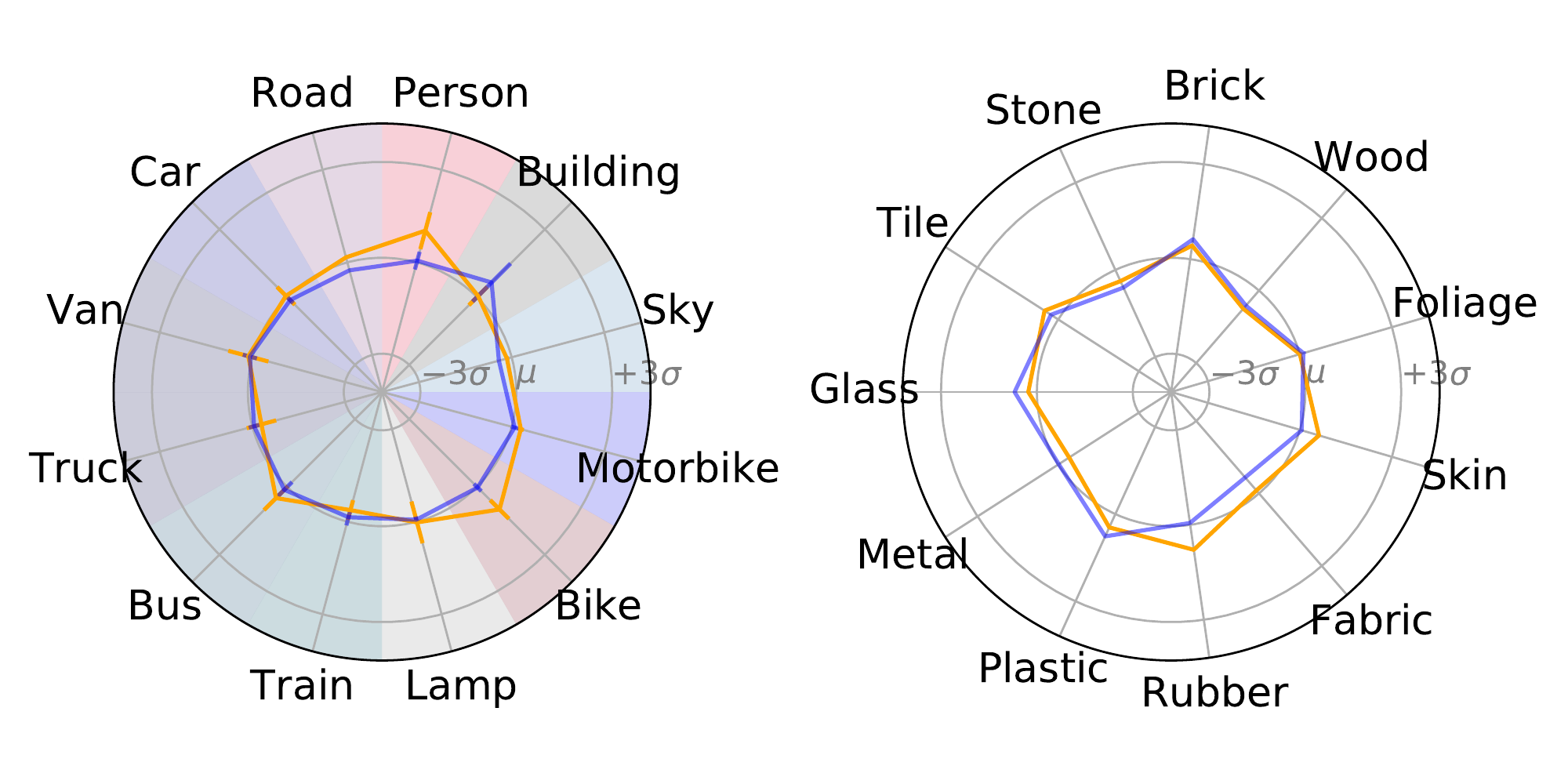}
        \end{center}
        \vspace{-10pt}
        \caption{Part and material activations}
    \end{subfigure}
    \caption{Selection of error examples from four different clusters. \textit{Blue:} SB activations used to classify the center pixel of the $3$x$3$ rectangle in white. \textit{Orange:} Average SB activations of the closest true positive cluster (counterfactual) of the target class for reference.}
    \label{fig:errors}
    \vspace{-20pt}
\end{figure}

For visualization purposes, we present only a selection of errors in figure~\ref{fig:errors} organized columnwise by the image-ROI, the groundtruth and predicted segmentation as well as the semantic concept activations for the $70$ concepts, split into parts and materials.
To simplify, we group the parts by their corresponding object category (as indicated in table~\ref{tab:selection_of_relevant}) and report mean and standard deviation within it.
By standardizing the SB-activations, on all samples from the Cityscapes validation set, to zero mean and unit variance we gain insight into the relative amplitude of each evidence.

Based on the clusters, we identify three patterns that result in misclassification:\\
\textbf{Misinterpretation} of image features. The first row of figure~\ref{fig:errors} originates from a cluster with $41$ samples and are all related to the misclassification of construction beacons (target class: \textit{sign}) as \textit{persons} with confidence of over $90\%$ (based on the softmax probabilitiy differences of the top $2$ logits).
The SB activations give insights into why that might be the case.
In particular, the material concept \textit{skin} is with over $4$ times the standard deviation highly activated, likely being caused by the color of the beacons.
We also observe that the part concept \textit{foot} is typically highly activated for these cases.
See figure~\ref{fig:ex_sign_as_person} in the appendix for full details on all part-concept activations.

\textbf{Conflicting evidence} which the SBN is not able to resolve properly.
We show here $2$ examples from $2$ clusters in the rows $2$ and $3$ in figure~\ref{fig:errors}, in which persons are either misclassified as car or as building.
Both with very high confidence.
In particular, in both cases there is clear evidence for the correct concepts, yet conflicting evidence diffuses from the immediate surroundings resulting in superseding evidence for the confused class.

\textbf{Missing evidence:} The last pattern we observe is lack of concept evidence, mostly in dark or blurry regions.
Though we find examples like the last row in figure~\ref{fig:errors} where the SB has hardly any evidence for the correct class even though it is clearly visible.
The example shown corresponds to a person standing in a door, the focus is on the legs of the person.
Here, the SB reveals that the detection for the concept \textit{leg}, \textit{torso} and \textit{foot} fails (see figure~\ref{fig:ex_building_as_person} in the appendix for individual part activations), indicating that the information might have been lost upstream.

\subsection{Confidence prediction}
\label{sec:exp:confidence}

As introduced in section \ref{sec:sbn:confidence}, we want to use the SB outputs to give an estimate of confidence for a given classification.
We proceed by using the SB@pyramid setup and train a binary classifier per class with the objective to distinguish between the class being present or not.
A sigmoid activation at the output results in a probability that is a direct proxy of confidence.
We train our classifiers on the coarse training dataset of Cityscapes, but filter errors made by our SBN (and potential labeling errors) to gather a dataset of only true positives and true negatives.
As architectures, we compare a single linear layer with a fully connected layer with one hidden layer of size $350$ and ReLU non-linearity, which are both trained the same way as the SBNs according to section~\ref{sec:sbn:implementation}.
Both have access to the same $3$x$3$ receptive field as the SBN classifier.

\begin{figure}[t]
\centering
\begin{subfigure}[t]{0.32\linewidth}
    \begin{center}
      \includegraphics[width=\linewidth]{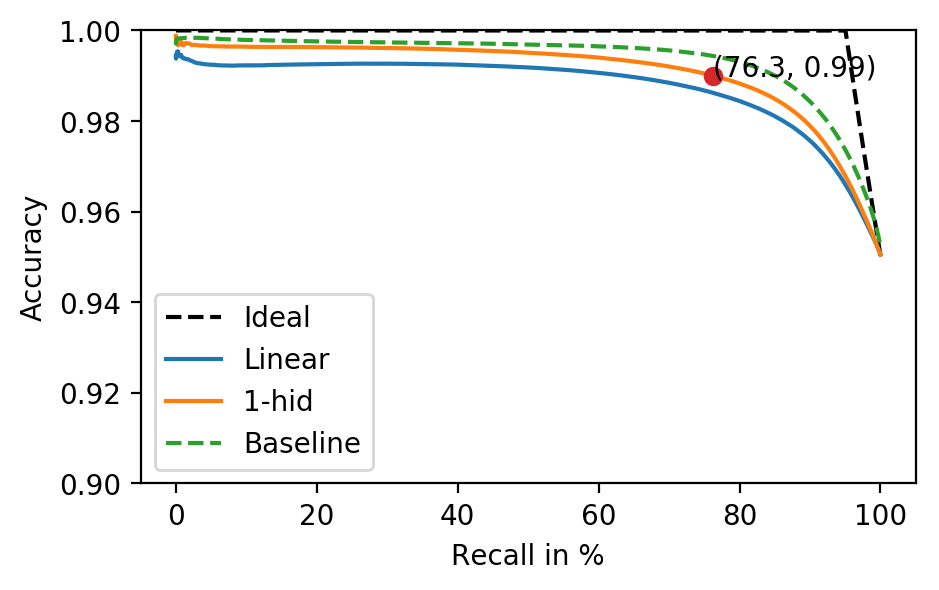}
    \end{center}
    \vspace{-10pt}
    \caption{For all classes. $76.3\%$ of all pixels can be assessed with $99\%$ accuracy.}
\end{subfigure}
\hfill
\begin{subfigure}[t]{0.32\linewidth}
    \begin{center}
      \includegraphics[width=\linewidth]{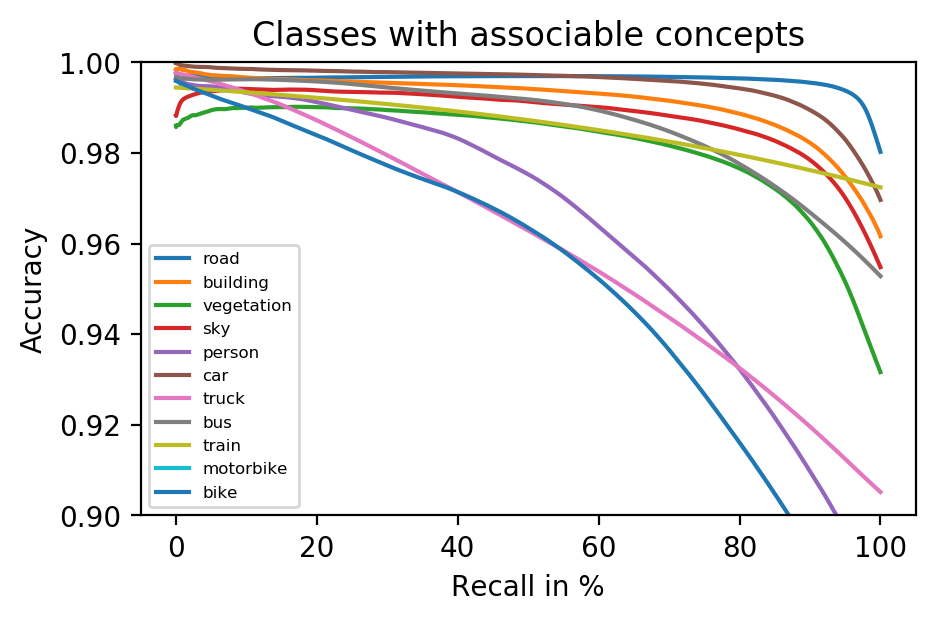}
    \end{center}
    \vspace{-10pt}
    \caption{For classes that have directly associable concepts (based on 1-hid model)}
    \label{fig:confidence_prediction:associable}
\end{subfigure}
\hfill
\begin{subfigure}[t]{0.32\linewidth}
    \begin{center}
      \includegraphics[width=\linewidth]{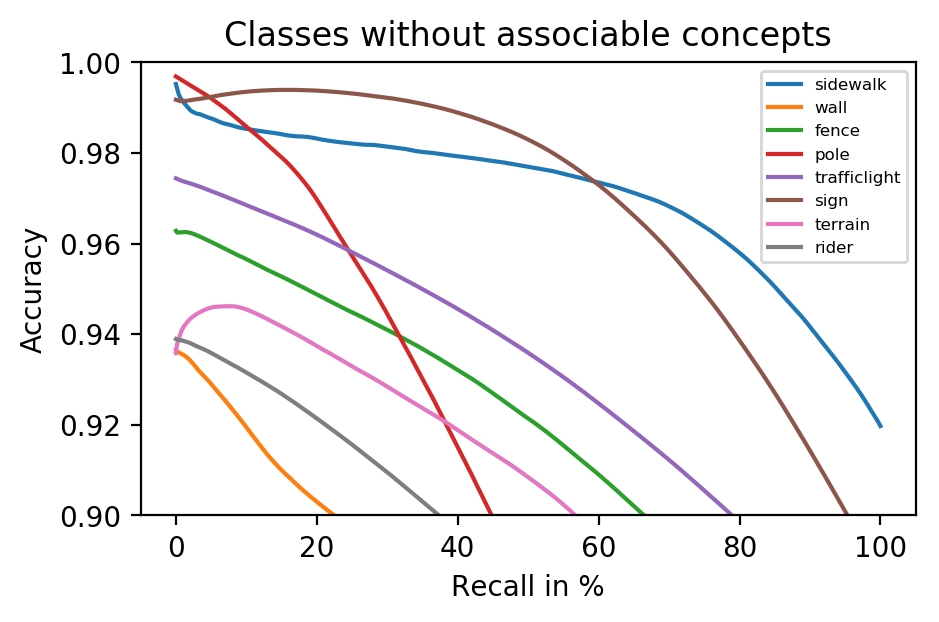}
    \end{center}
    \vspace{-10pt}
    \caption{For classes that do not have directly associable concepts (based on 1-hid model)}
    \label{fig:confidence_prediction:notassociable}
\end{subfigure}
\caption{Accuracy assessment of the networks predictions with our proposed confidence metric. The x-axis states the number of pixels considered, where all pixels are ordered by their respective confidence score and the y-axis states the accuracy of bespoken pixels. Thus, the ideal case is a straight line with perfect accuracy until all misclassified pixels are evaluated last resulting in a linear drop. All metrics are evaluated on the SB after the PSPNet-pyramid.}
\label{fig:confidence_prediction}
\end{figure}

To evaluate the effectiveness of these confidence predictors, we let the SBN classify a given pixel and subsequently look up the confidence estimate for the predicted class.
Doing this for all images in the Cityscapes validation set, we rank the pixels from highest to lowest confidence and plot them with respect to their accuracy in figure~\ref{fig:confidence_prediction}, but smoothen the curve with an average filter of length $10,000$ to reduce the impact of few errors on small sample sizes.
As baseline, we use the difference between the largest two softmax probabilities of the SBN to estimate an alternative confidence for each pixel.

We observe that our confidence predictions based on the $1$-hidden layer network is only marginally worse than the non-interpretable baseline. We find that classes that have associable concepts in our SB show a more accurate confidence prediction -- that in turns results in a better ranking (compare figure~\ref{fig:confidence_prediction:associable} and~\ref{fig:confidence_prediction:notassociable}). 
Remarkably, our interpretable model can predict $76.3\%$ of all pixels with an accuracy of $99\%$.

\section{Discussion \& Conclusion}

We have introduced the concept of SBs for street scene segmentation as a means to take a leap towards inherently interpretable models that do not impair performance.
We have shown how simple transformations of high level features into an interpretable representation space can drastically simplify the process of understanding the errors the network makes, can enable the probing of the networks predictions for the presence or absence of concept evidence in the SB and can approximate the final non linear classification with a linear classifier to give an assessment on the accuracy of the classification.

Concluding, SBN can add helpful information on the inner workings of the decision making process.
Yet, the power of the SBN can be increased further by placing multiple SBs at various layers allowing the inspection of vanishing concept evidences.

\section*{Acknowledgments}
This research was supported by the Bosch Computer Vision Research Lab Hildesheim, Germany.
We thank Dimitrios Bariamis and Oliver Lange for the insightful discussions.

\newpage
{\small
\bibliography{egbib}

\begin{thebibliography}{10}

\bibitem{al2017contextual}
Maruan Al-Shedivat, Avinava Dubey, and Eric~P Xing.
\newblock Contextual explanation networks.
\newblock {\em arXiv:1705.10301}, 2017.

\bibitem{bach2015pixel}
Sebastian Bach, Alexander Binder, Gr{\'e}goire Montavon, Frederick Klauschen,
  Klaus-Robert M{\"u}ller, and Wojciech Samek.
\newblock On pixel-wise explanations for non-linear classifier decisions by
  layer-wise relevance propagation.
\newblock {\em PloS one}, 10(7):e0130140, 2015.

\bibitem{bau2017network}
David Bau, Bolei Zhou, Aditya Khosla, Aude Oliva, and Antonio Torralba.
\newblock Network dissection: Quantifying interpretability of deep visual
  representations.
\newblock In {\em CVPR}, 2017.

\bibitem{bay2006surf}
Herbert Bay, Tinne Tuytelaars, and Luc Van~Gool.
\newblock Surf: Speeded up robust features.
\newblock In {\em ECCV}, 2006.

\bibitem{bell2013opensurfaces}
Sean Bell, Paul Upchurch, Noah Snavely, and Kavita Bala.
\newblock Opensurfaces: A richly annotated catalog of surface appearance.
\newblock {\em ACM Transactions on Graphics (TOG)}, 32(4):111, 2013.

\bibitem{chen2018deeplab}
Liang-Chieh Chen, George Papandreou, Iasonas Kokkinos, Kevin Murphy, and Alan~L
  Yuille.
\newblock Deeplab: Semantic image segmentation with deep convolutional nets,
  atrous convolution, and fully connected crfs.
\newblock {\em TPAMI}, 40(4):834--848, 2018.

\bibitem{chen2014detect}
Xianjie Chen, Roozbeh Mottaghi, Xiaobai Liu, Sanja Fidler, Raquel Urtasun, and
  Alan Yuille.
\newblock Detect what you can: Detecting and representing objects using
  holistic models and body parts.
\newblock In {\em CVPR}, 2014.

\bibitem{cordts2016cityscapes}
Marius Cordts, Mohamed Omran, Sebastian Ramos, Timo Rehfeld, Markus Enzweiler,
  Rodrigo Benenson, Uwe Franke, Stefan Roth, and Bernt Schiele.
\newblock The cityscapes dataset for semantic urban scene understanding.
\newblock In {\em CVPR}, 2016.

\bibitem{goodfellow2014explaining}
Ian~J Goodfellow, Jonathon Shlens, and Christian Szegedy.
\newblock Explaining and harnessing adversarial examples.
\newblock {\em arXiv:1412.6572}, 2014.

\bibitem{he2016deep}
Kaiming He, Xiangyu Zhang, Shaoqing Ren, and Jian Sun.
\newblock Deep residual learning for image recognition.
\newblock In {\em CVPR}, 2016.

\bibitem{ward1963}
Joe H.~Ward Jr.
\newblock Hierarchical grouping to optimize an objective function.
\newblock {\em Journal of the American Statistical Association},
  58(301):236--244, 1963.

\bibitem{kim2018interpretability}
Been Kim, Martin Wattenberg, Justin Gilmer, Carrie Cai, James Wexler, Fernanda
  Viegas, et~al.
\newblock Interpretability beyond feature attribution: Quantitative testing
  with concept activation vectors (tcav).
\newblock In {\em ICML}, 2018.

\bibitem{kindermans2017reliability}
Pieter-Jan Kindermans, Sara Hooker, Julius Adebayo, Maximilian Alber, Kristof~T
  Sch{\"u}tt, Sven D{\"a}hne, Dumitru Erhan, and Been Kim.
\newblock The (un) reliability of saliency methods.
\newblock {\em arXiv:1711.00867}, 2017.

\bibitem{kurakin2016adversarial}
Alexey Kurakin, Ian Goodfellow, and Samy Bengio.
\newblock Adversarial examples in the physical world.
\newblock {\em arXiv:1607.02533}, 2016.

\bibitem{li2010object}
Li-Jia Li, Hao Su, Li~Fei-Fei, and Eric~P Xing.
\newblock Object bank: A high-level image representation for scene
  classification \& semantic feature sparsification.
\newblock In {\em NIPS}, 2010.

\bibitem{li2018deep}
Oscar Li, Hao Liu, Chaofan Chen, and Cynthia Rudin.
\newblock Deep learning for case-based reasoning through prototypes: A neural
  network that explains its predictions.
\newblock In {\em AAAI}, 2018.

\bibitem{lipton2018mythos}
Zachary~C Lipton.
\newblock The mythos of model interpretability.
\newblock {\em Queue}, 16(3):30, 2018.

\bibitem{lowe2004distinctive}
David~G Lowe.
\newblock Distinctive image features from scale-invariant keypoints.
\newblock {\em IJCV}, 60(2):91--110, 2004.

\bibitem{madry2017towards}
Aleksander Madry, Aleksandar Makelov, Ludwig Schmidt, Dimitris Tsipras, and
  Adrian Vladu.
\newblock Towards deep learning models resistant to adversarial attacks.
\newblock In {\em ICLR}, 2018.

\bibitem{melis2018towards}
David~Alvarez Melis and Tommi Jaakkola.
\newblock Towards robust interpretability with self-explaining neural networks.
\newblock In {\em NIPS}, 2018.

\bibitem{selvaraju2017grad}
Ramprasaath~R Selvaraju, Michael Cogswell, Abhishek Das, Ramakrishna Vedantam,
  Devi Parikh, Dhruv Batra, et~al.
\newblock Grad-cam: Visual explanations from deep networks via gradient-based
  localization.
\newblock In {\em ICCV}, pages 618--626, 2017.

\bibitem{shrikumar2017learning}
Avanti Shrikumar, Peyton Greenside, and Anshul Kundaje.
\newblock Learning important features through propagating activation
  differences.
\newblock In {\em ICML}, 2017.

\bibitem{simonyan2013deep}
Karen Simonyan, Andrea Vedaldi, and Andrew Zisserman.
\newblock Deep inside convolutional networks: Visualising image classification
  models and saliency maps.
\newblock {\em arXiv:1312.6034}, 2013.

\bibitem{sundararajan2017axiomatic}
Mukund Sundararajan, Ankur Taly, and Qiqi Yan.
\newblock Axiomatic attribution for deep networks.
\newblock In {\em ICML}, 2017.

\bibitem{szegedy2013intriguing}
Christian Szegedy, Wojciech Zaremba, Ilya Sutskever, Joan Bruna, Dumitru Erhan,
  Ian Goodfellow, and Rob Fergus.
\newblock Intriguing properties of neural networks.
\newblock {\em arXiv:1312.6199}, 2013.

\bibitem{wachter2017counterfactual}
Sandra Wachter, Brent Mittelstadt, and Chris Russell.
\newblock Counterfactual explanations without opening the black box: Automated
  decisions and the gdpr.
\newblock {\em Harvard Journal of Law \& Technology}, 31(2):2018, 2017.

\bibitem{xiao2018unified}
Tete Xiao, Yingcheng Liu, Bolei Zhou, Yuning Jiang, and Jian Sun.
\newblock Unified perceptual parsing for scene understanding.
\newblock In {\em ECCV}, 2018.

\bibitem{yosinski2015understanding}
Jason Yosinski, Jeff Clune, Anh Nguyen, Thomas Fuchs, and Hod Lipson.
\newblock Understanding neural networks through deep visualization.
\newblock {\em arXiv:1506.06579}, 2015.

\bibitem{zeiler2014visualizing}
Matthew~D Zeiler and Rob Fergus.
\newblock Visualizing and understanding convolutional networks.
\newblock In {\em ECCV}, 2014.

\bibitem{zhao2017pyramid}
Hengshuang Zhao, Jianping Shi, Xiaojuan Qi, Xiaogang Wang, and Jiaya Jia.
\newblock Pyramid scene parsing network.
\newblock In {\em CVPR}, 2017.

\bibitem{Zhou2015ObjectDE}
Bolei Zhou, Aditya Khosla, {\`A}gata Lapedriza, Aude Oliva, and Antonio
  Torralba.
\newblock Object detectors emerge in deep scene cnns.
\newblock {\em CoRR}, 2015.

\bibitem{zhou2017scene}
Bolei Zhou, Hang Zhao, Xavier Puig, Sanja Fidler, Adela Barriuso, and Antonio
  Torralba.
\newblock Scene parsing through ade20k dataset.
\newblock In {\em CVPR}, 2017.

\bibitem{zintgraf2017visualizing}
Luisa~M Zintgraf, Taco~S Cohen, Tameem Adel, and Max Welling.
\newblock Visualizing deep neural network decisions: Prediction difference
  analysis.
\newblock {\em arXiv:1702.04595}, 2017.

\end{thebibliography}
\bibliographystyle{plain}
}

\newpage
\appendix
\renewcommand\thefigure{\thesection.\arabic{figure}}
\renewcommand\thetable{\thesection.\arabic{table}}
{\Large \textbf{Supplementary Material}}
\section{Intro}

This material contains additional information that otherwise would not have fit in the main paper.
It is organized in three parts.
The selection of concepts from the \textit{Broden+} dataset that we think are relevant to the task of street scene segmentation and the Cityscapes classes are listed in section~\ref{app:concepts}.

Additional implementation details on our agglomerative clustering algorithm is presented in section~\ref{app:clustering}.

Finally, we show more examples of failure cases at the end of this document.

\section{Selection of concepts for Cityscapes}
\label{app:concepts}

\begin{table}[h]
    \centering
    \begin{footnotesize}
    \begin{tabular}{r|r|l}
        Materials & \multicolumn{2}{l}{Brick,  Fabric,  Foliage,  Glass,  Metal,  Plastic,  Rubber, Skin,  Stone,  Tile,  Wood} \\\hline
        \multirow{12}{*}{Parts} & Sky & Cloud\\
        & Building & Window, Door, Roof, Shop, Wall\\
        & Person & Leg, Head, Torso, Arm Eye, Ear, Nose, Hand, Hair, Mouth, Foot, Eyebrow, Back\\
        & Road & Crosswalk\\
        & Car & Window, Door, Wheel, Headlight, Mirror, Roof, Taillight, Windshield, Bumber\\
        & Van & Window, Door, Wheel, Headlight, Taillight, Windshield\\
        & Truck & Wheel, Windshield\\
        & Bus & Window, Door, Wheel, Headlight, Mirror\\
        & Train & Head, Headlight, Headroof, Coachroof\\
        & Lamp & Arm, Shade, Bulb\\
        & Bike & Wheel, Handle, Saddle, Chain\\
        & Motorbike & Wheel, Headlight, Handle\\
    \end{tabular}
    \end{footnotesize}
    \caption{Selection of $70$ concepts in total that we deemed relevant for the task of segmentation}
    \label{tab:concepts}
\end{table}

\section{Implementation details - Clustering}
\label{app:clustering}

To populate the semantic space in preparation for the clustering, we source samples from the Cityscapes validation set with $500$ images and the coarse training set with $20,000$ images, but remove small error pixels and borders via dilation and erosion on both the prediction and the groundtruth masks.

We eventually find our error clusters by letting the algorithm build the complete tree and traverse it in a second step from top to bottom until we find clusters that satisfy our stopping criteria.
Here, we observe that using outcome type purity and maximal cluster size with the ward linkage method~\cite{ward1963} results subjectively in the most coherent error clusters and found $0.9$ for purity and $150$ for max. cluster size to work best.
To get an impression on the number and size of clusters we find with our setup, see table~\ref{tab:cluster_stats}, which reports statistics for three different classes.

\begin{table}
\centering
\begin{tabular}{r|c|c|c}
                                         & person  & car     & bike    \\\hline
Total number of samples                  & 236,346 & 369,136 & 215,640 \\
Number of error-samples                  & 36,326  & 169,116 & 39,240  \\
\#error-clusters of size \textgreater{}2 & 1105    & 2833    & 855     \\
accounting for fraction of samples       & 0.969   & 0.998   & 0.979   \\
Avg \#samples per cluster                & 31.9    & 59.1    & 45.1    \\
Avg. \#images per cluster                & 1.3     & 1.4     & 1.3    
\end{tabular}
\caption{Cluster statistics for three Cityscapes classes.}
\label{tab:cluster_stats}
\end{table}

\section{Supplementary error examples}

\begin{figure}
    \centering
    \begin{subfigure}[t]{0.59\linewidth}
        \begin{center}
           \includegraphics[width=\linewidth]{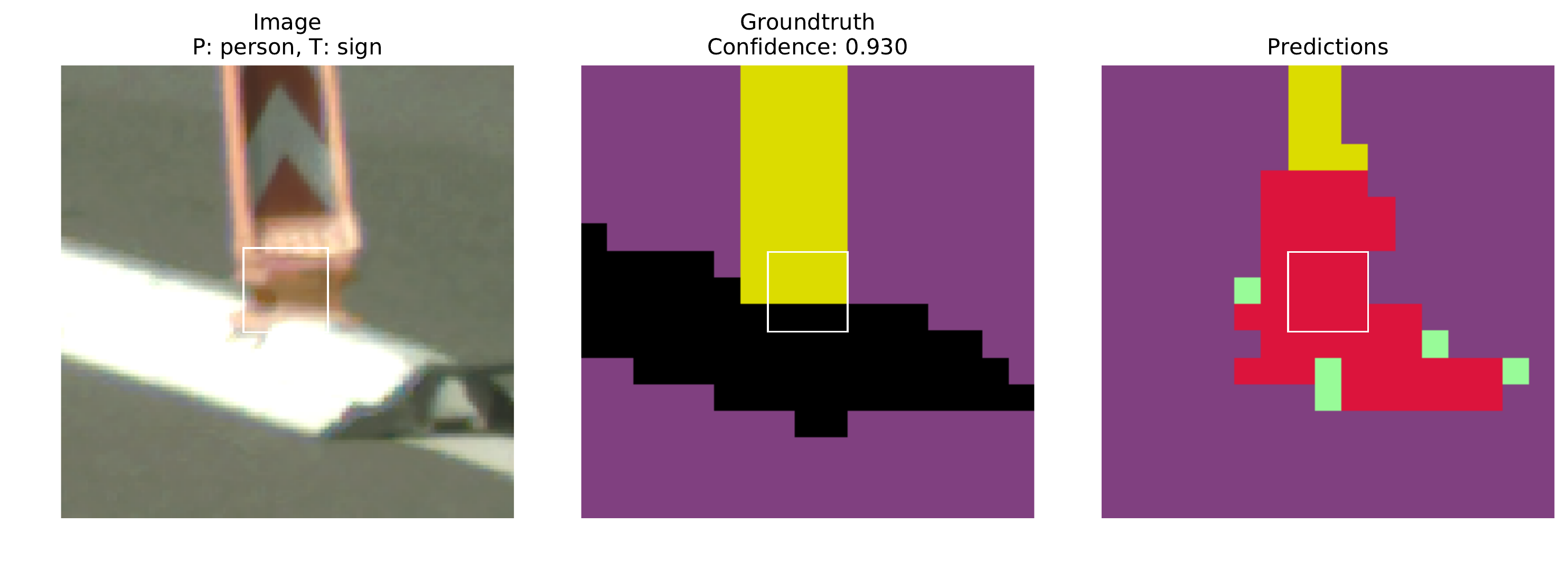}
        \end{center}
    \end{subfigure}
    \begin{subfigure}[t]{0.4\linewidth}
        \begin{center}
           \includegraphics[width=\linewidth]{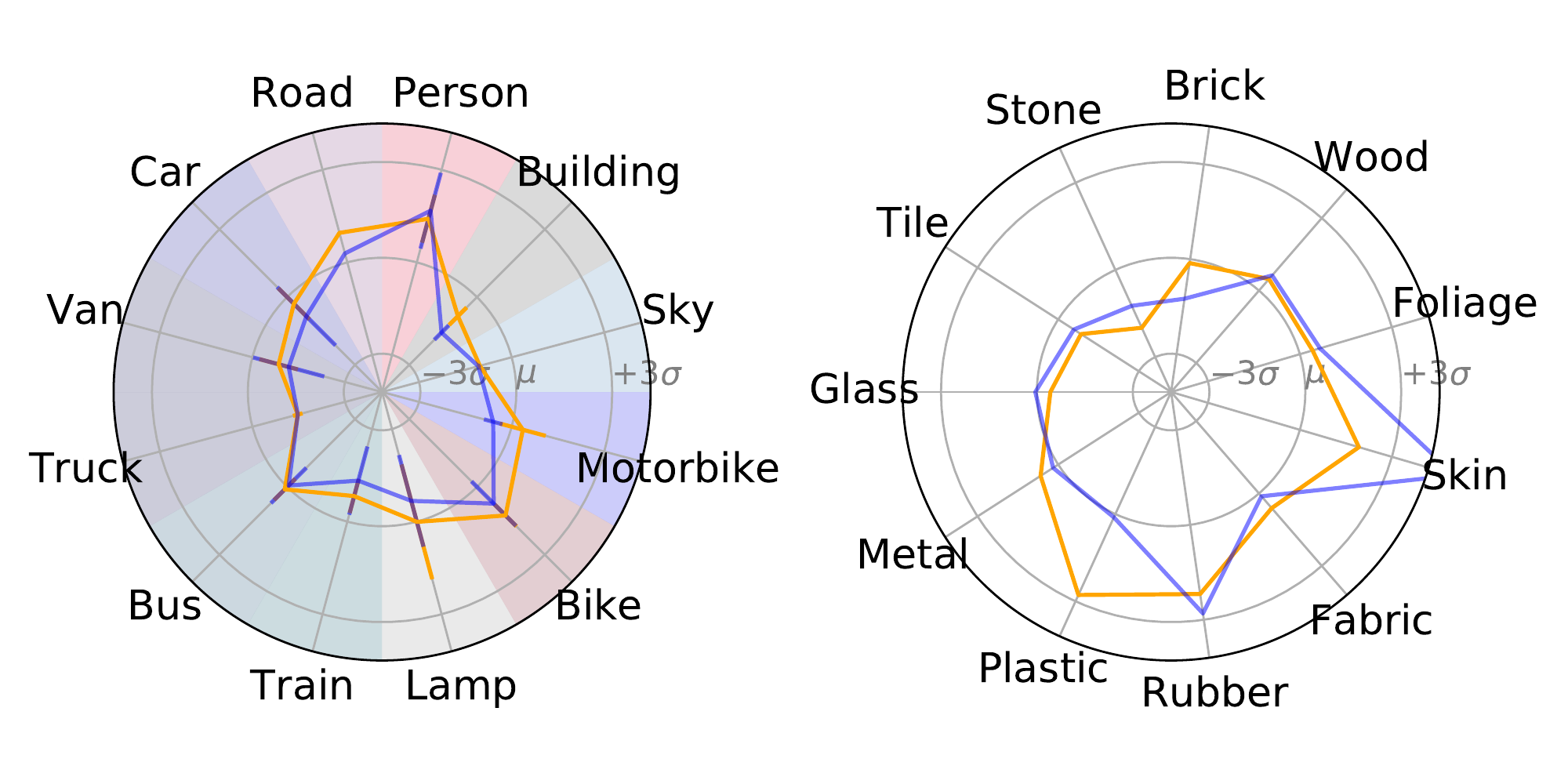}
        \end{center}
    \end{subfigure}
    
    \begin{subfigure}[t]{0.4\linewidth}
        \begin{center}
           \includegraphics[width=\linewidth]{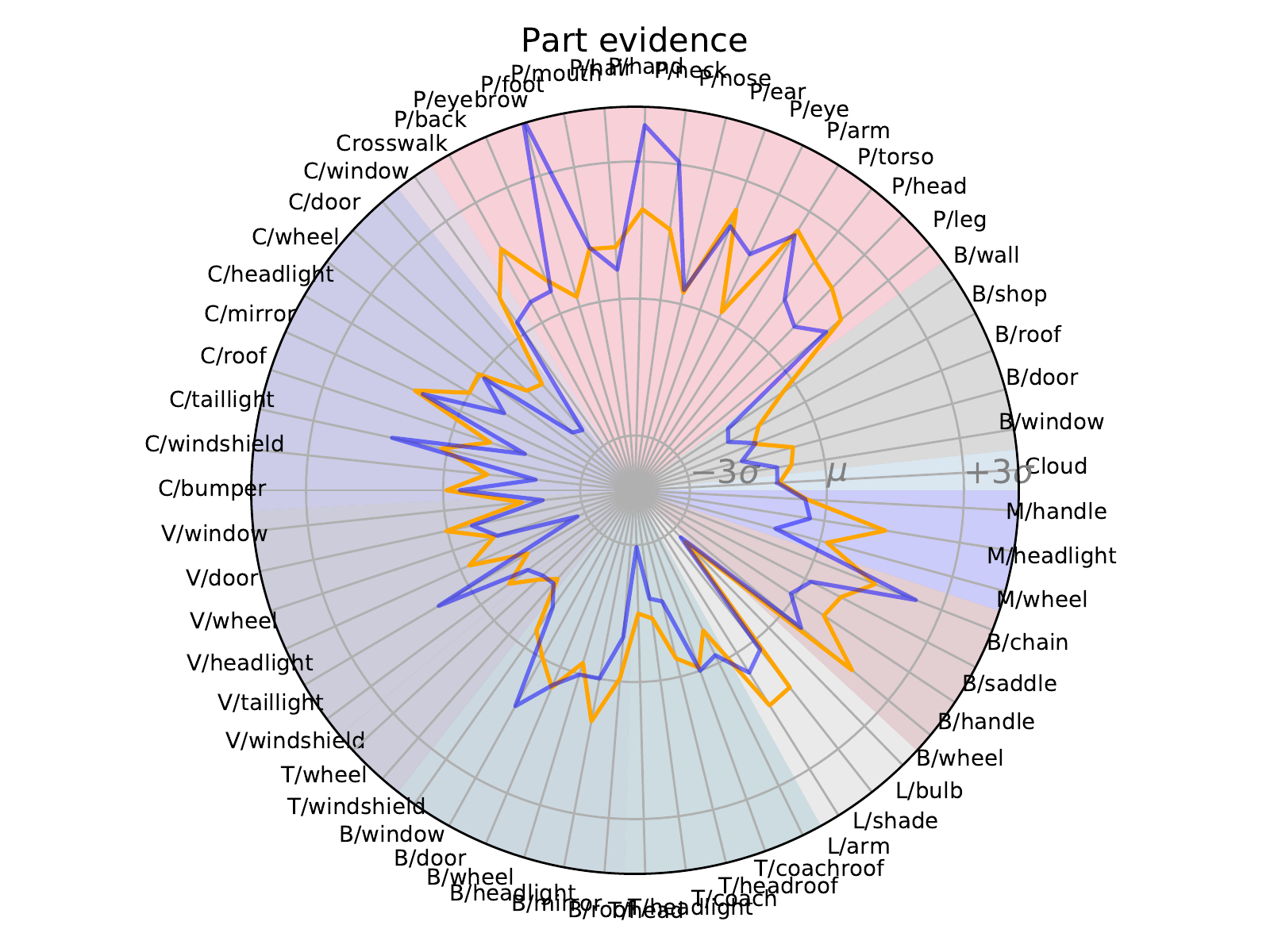}
        \end{center}
        \vspace{-15pt}
        \caption{Individual part activations}
    \end{subfigure}
    
    \begin{subfigure}[t]{0.59\linewidth}
        \begin{center}
           \includegraphics[width=\linewidth]{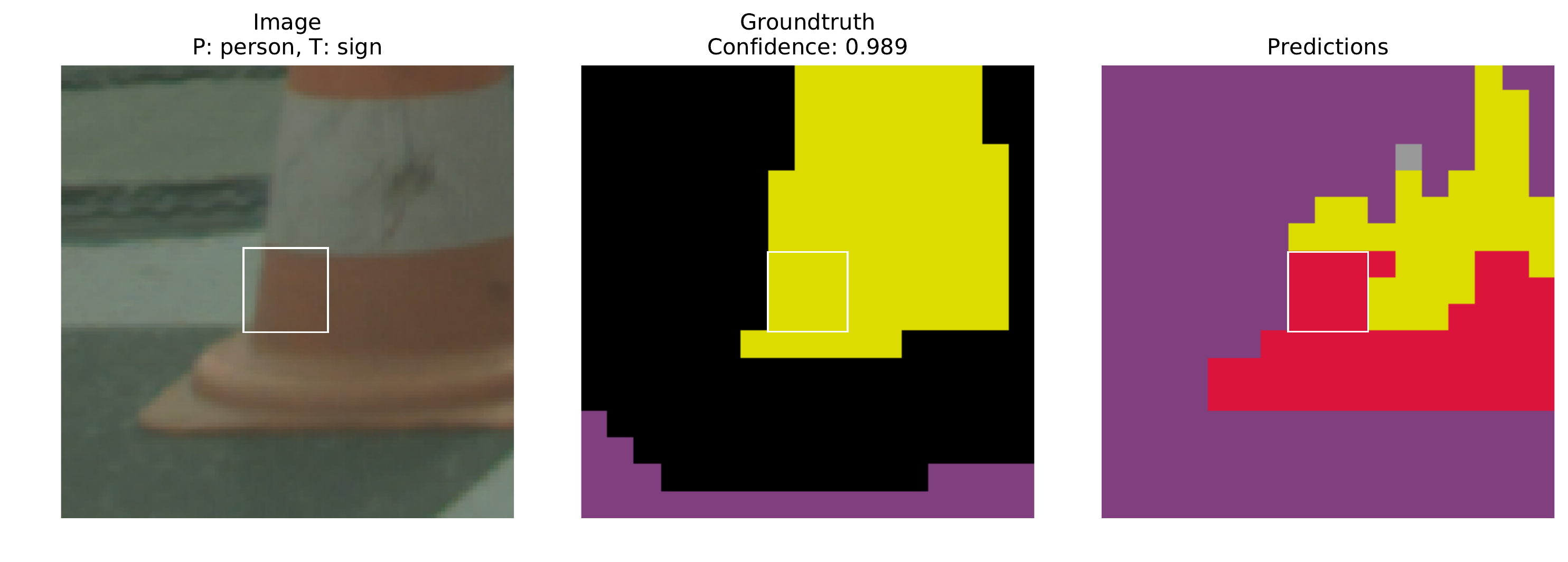}
        \end{center}
    \end{subfigure}
    \begin{subfigure}[t]{0.4\linewidth}
        \begin{center}
           \includegraphics[width=\linewidth]{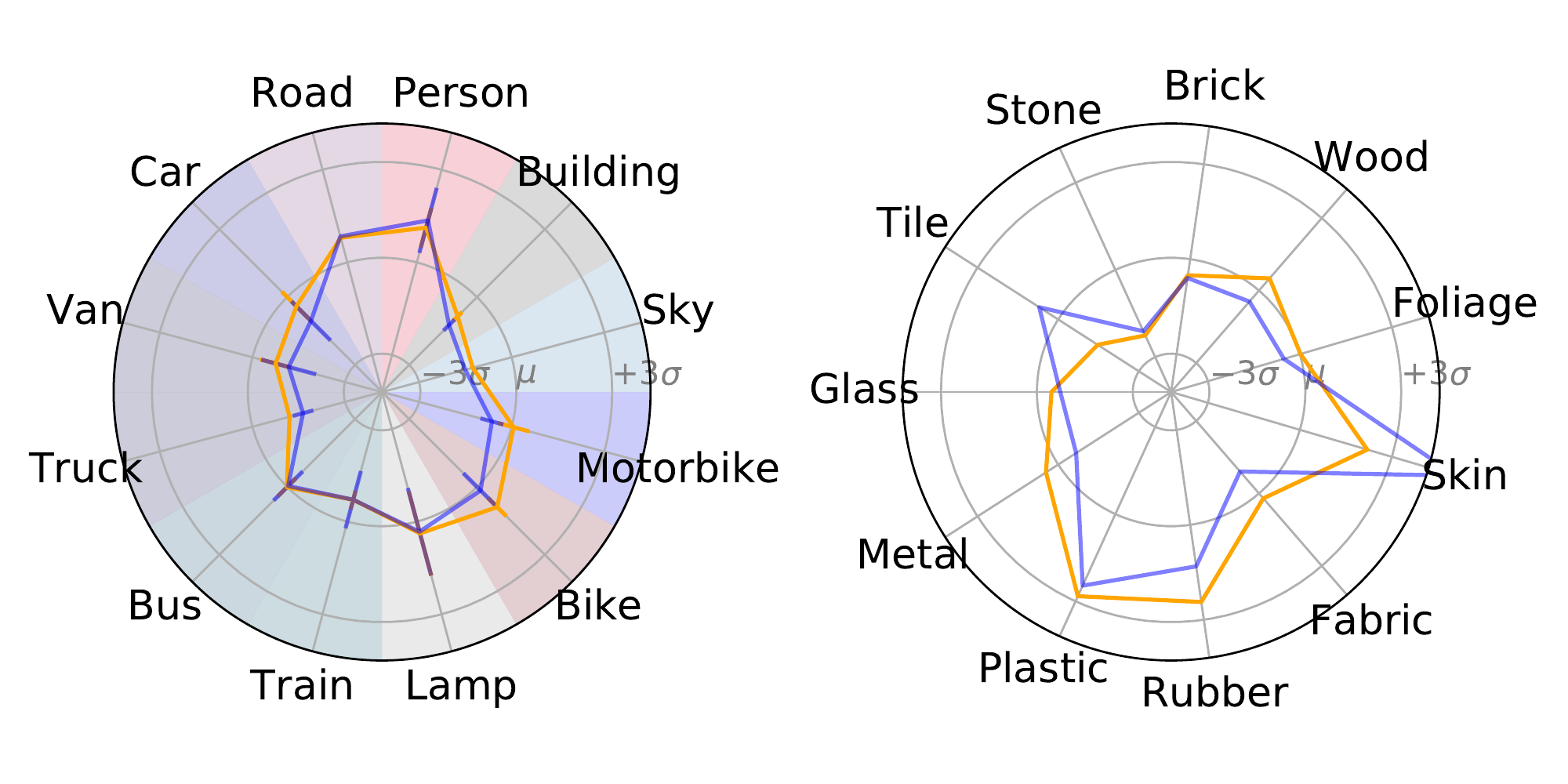}
        \end{center}
    \end{subfigure}
    
    \begin{subfigure}[t]{0.4\linewidth}
        \begin{center}
           \includegraphics[width=\linewidth]{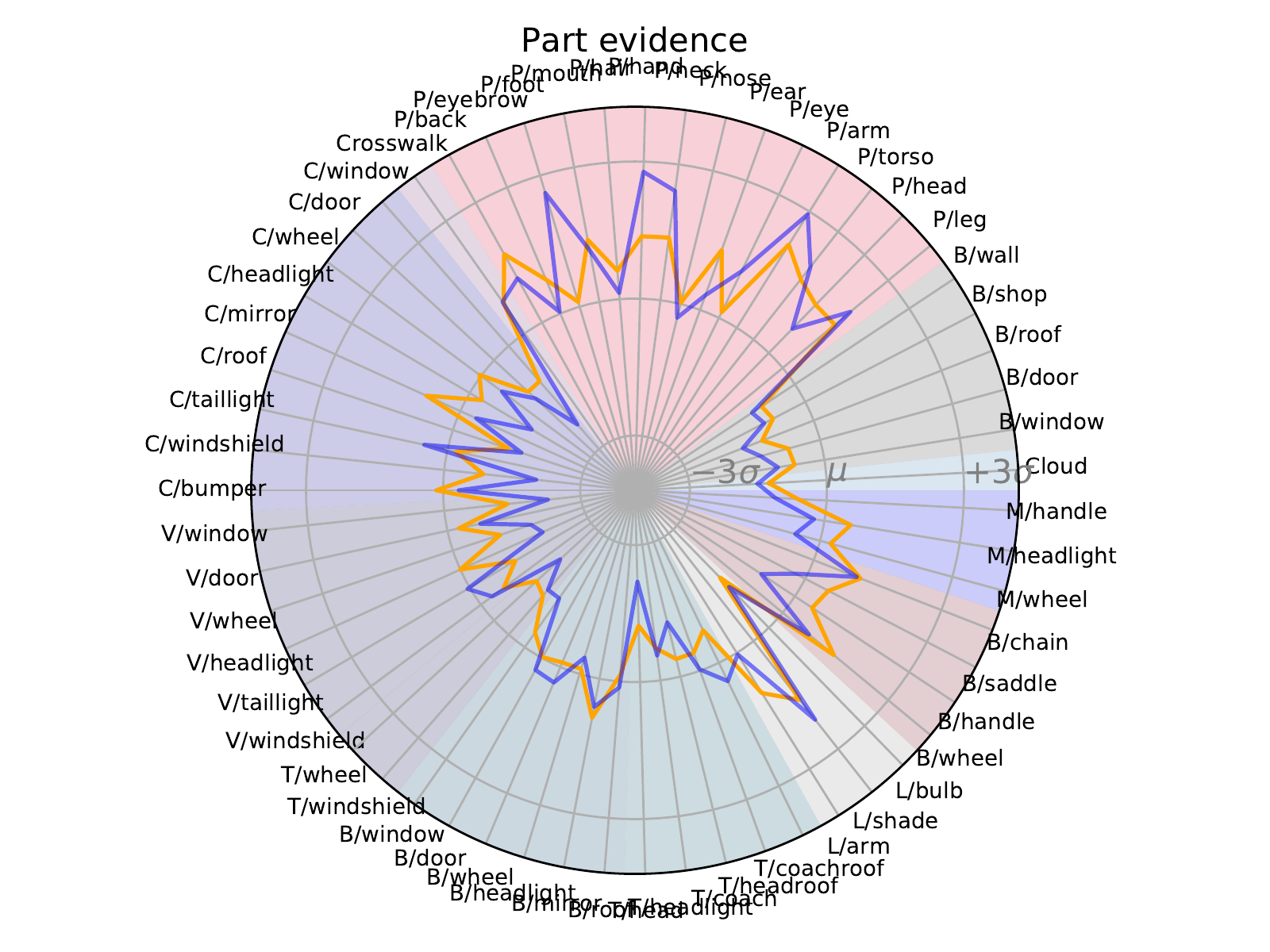}
        \end{center}
        \vspace{-15pt}
        \caption{Individual part activations}
    \end{subfigure}
    
    \begin{subfigure}[t]{0.59\linewidth}
        \begin{center}
           \includegraphics[width=\linewidth]{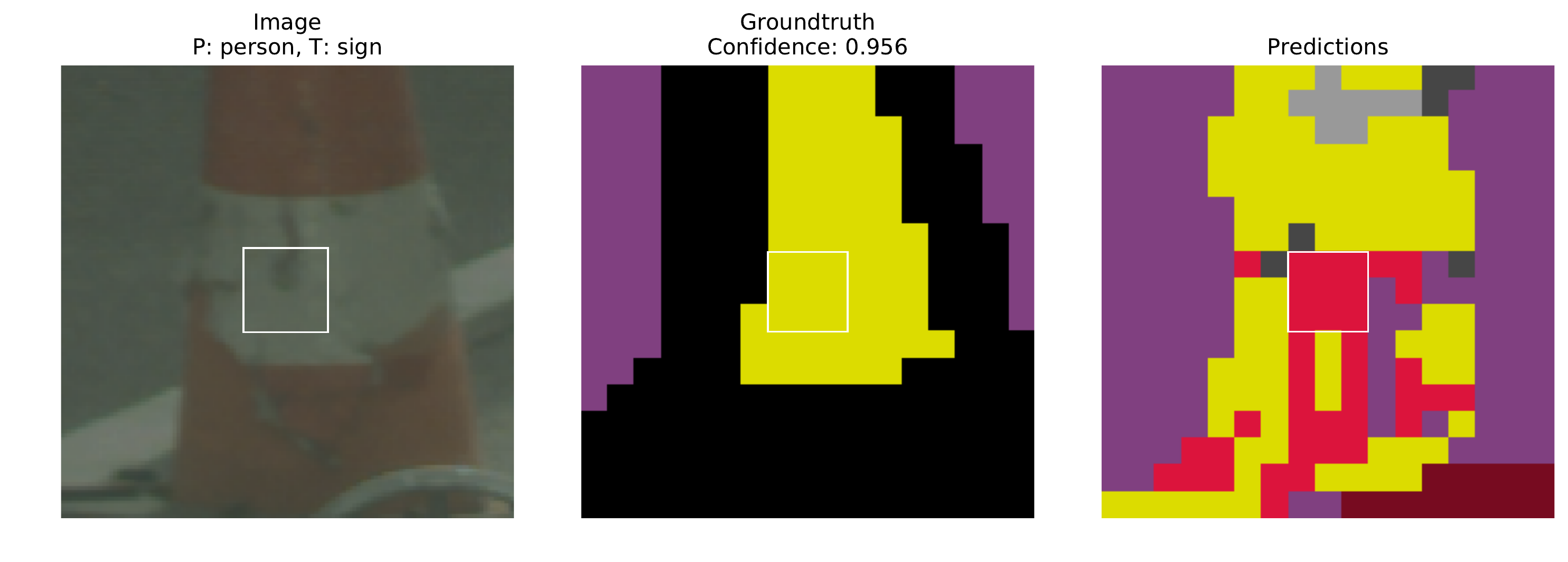}
        \end{center}
    \end{subfigure}
    \begin{subfigure}[t]{0.4\linewidth}
        \begin{center}
           \includegraphics[width=\linewidth]{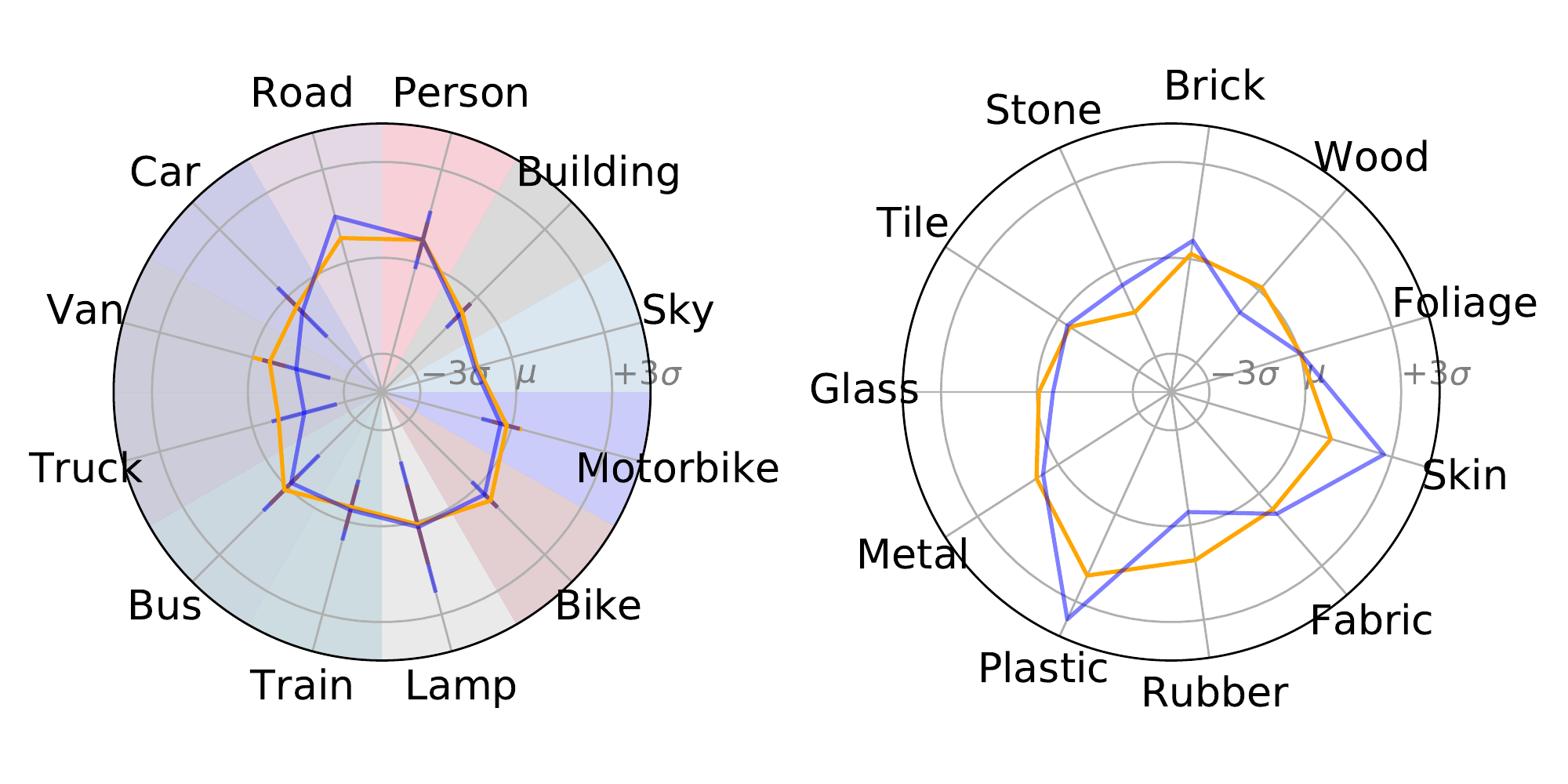}
        \end{center}
    \end{subfigure}
    
    \begin{subfigure}[t]{0.4\linewidth}
        \begin{center}
           \includegraphics[width=\linewidth]{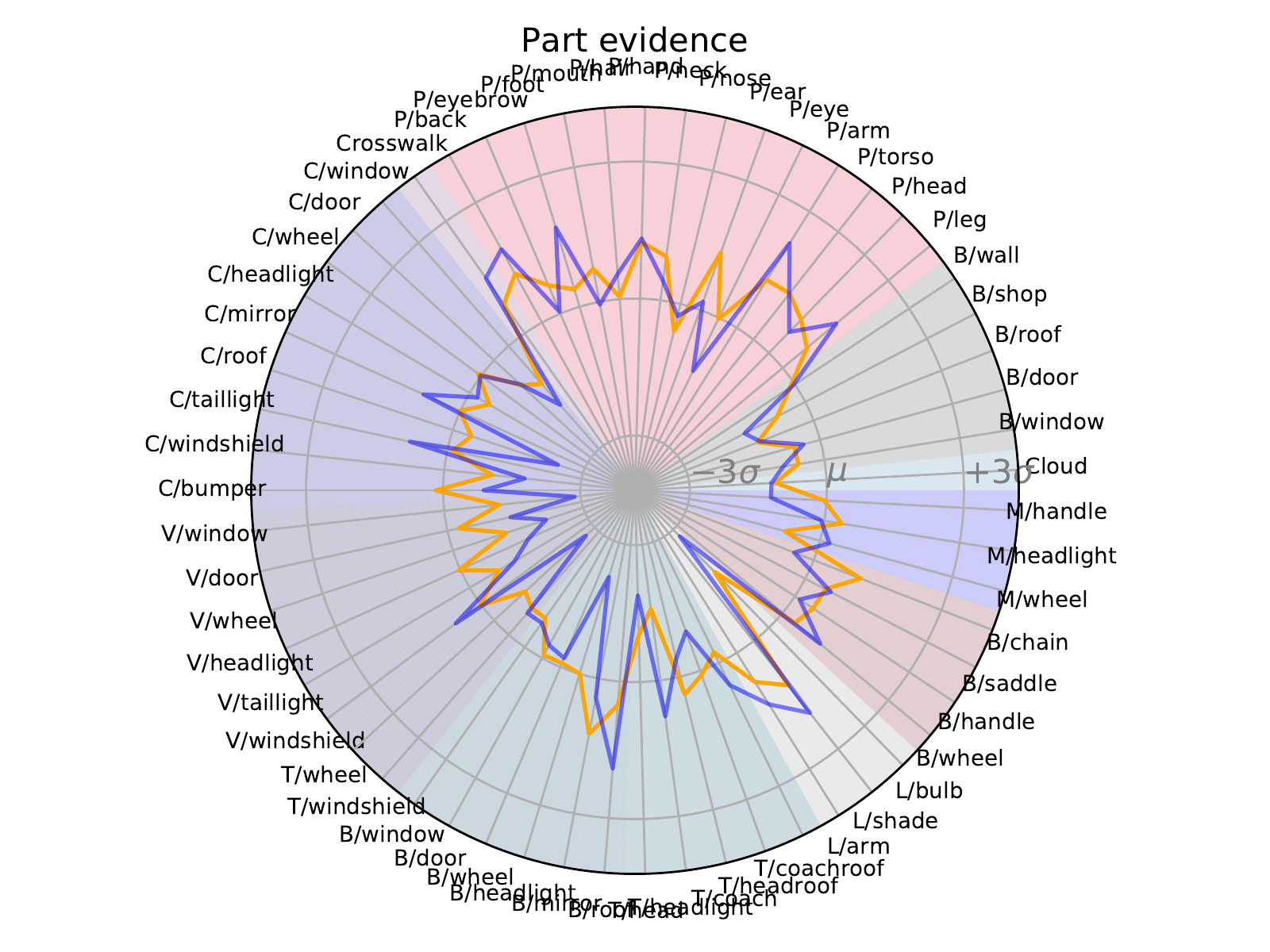}
        \end{center}
        \vspace{-15pt}
        \caption{Individual part activations}
    \end{subfigure}
    \caption{Selection of $3$ samples from a cluster of size $41$ related to error pixels in $3$ different images.}
    \label{fig:ex_sign_as_person}
\end{figure}

\begin{figure}
    \centering
    \begin{subfigure}[t]{0.59\linewidth}
        \begin{center}
           \includegraphics[width=\linewidth]{figures/error_cluster/1405_fns_person_as_car/0013_img_gt_pred.pdf}
        \end{center}
    \end{subfigure}
    \begin{subfigure}[t]{0.4\linewidth}
        \begin{center}
           \includegraphics[width=\linewidth]{figures/error_cluster/1405_fns_person_as_car/0013_part_and_mat.pdf}
        \end{center}
    \end{subfigure}
    
    \begin{subfigure}[t]{0.4\linewidth}
        \begin{center}
           \includegraphics[width=\linewidth]{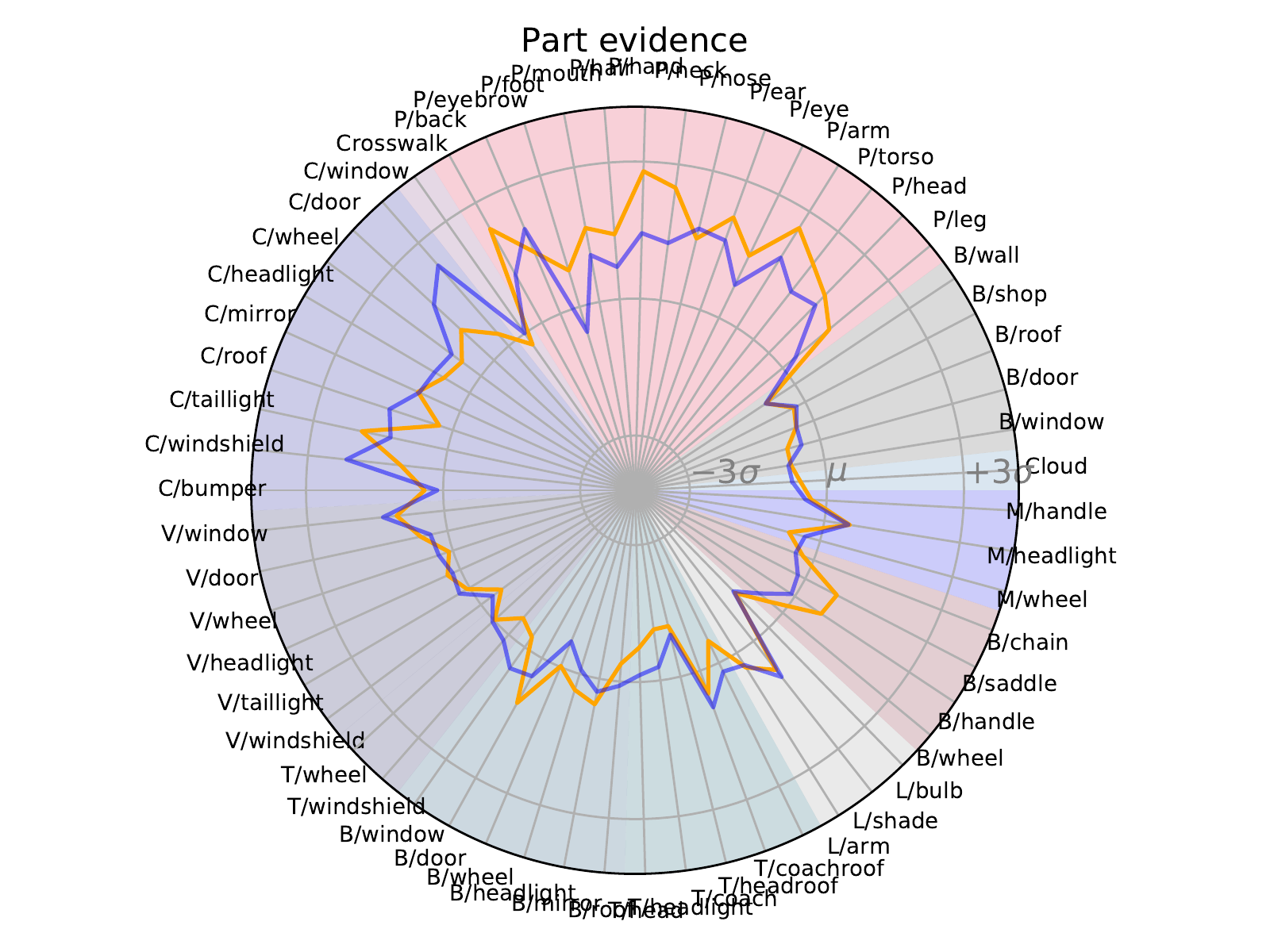}
        \end{center}
        \vspace{-15pt}
        \caption{Individual part activations}
    \end{subfigure}
    
    \begin{subfigure}[t]{0.59\linewidth}
        \begin{center}
           \includegraphics[width=\linewidth]{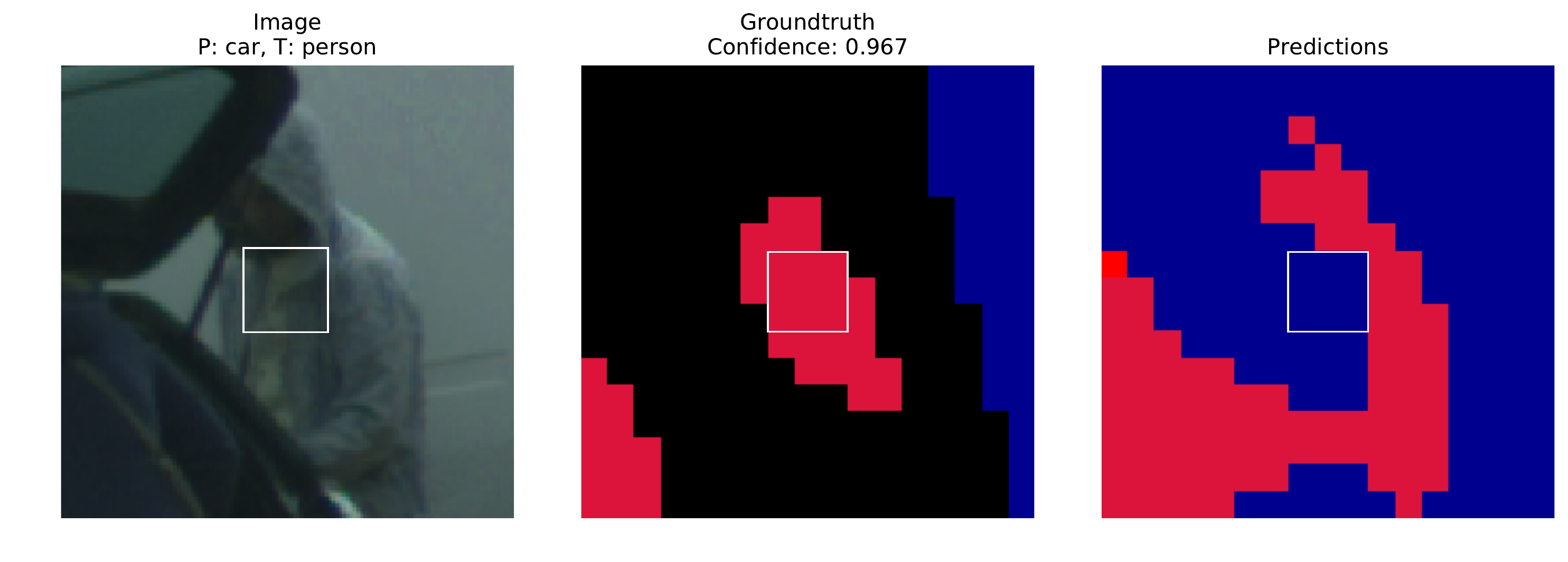}
        \end{center}
    \end{subfigure}
    \begin{subfigure}[t]{0.4\linewidth}
        \begin{center}
           \includegraphics[width=\linewidth]{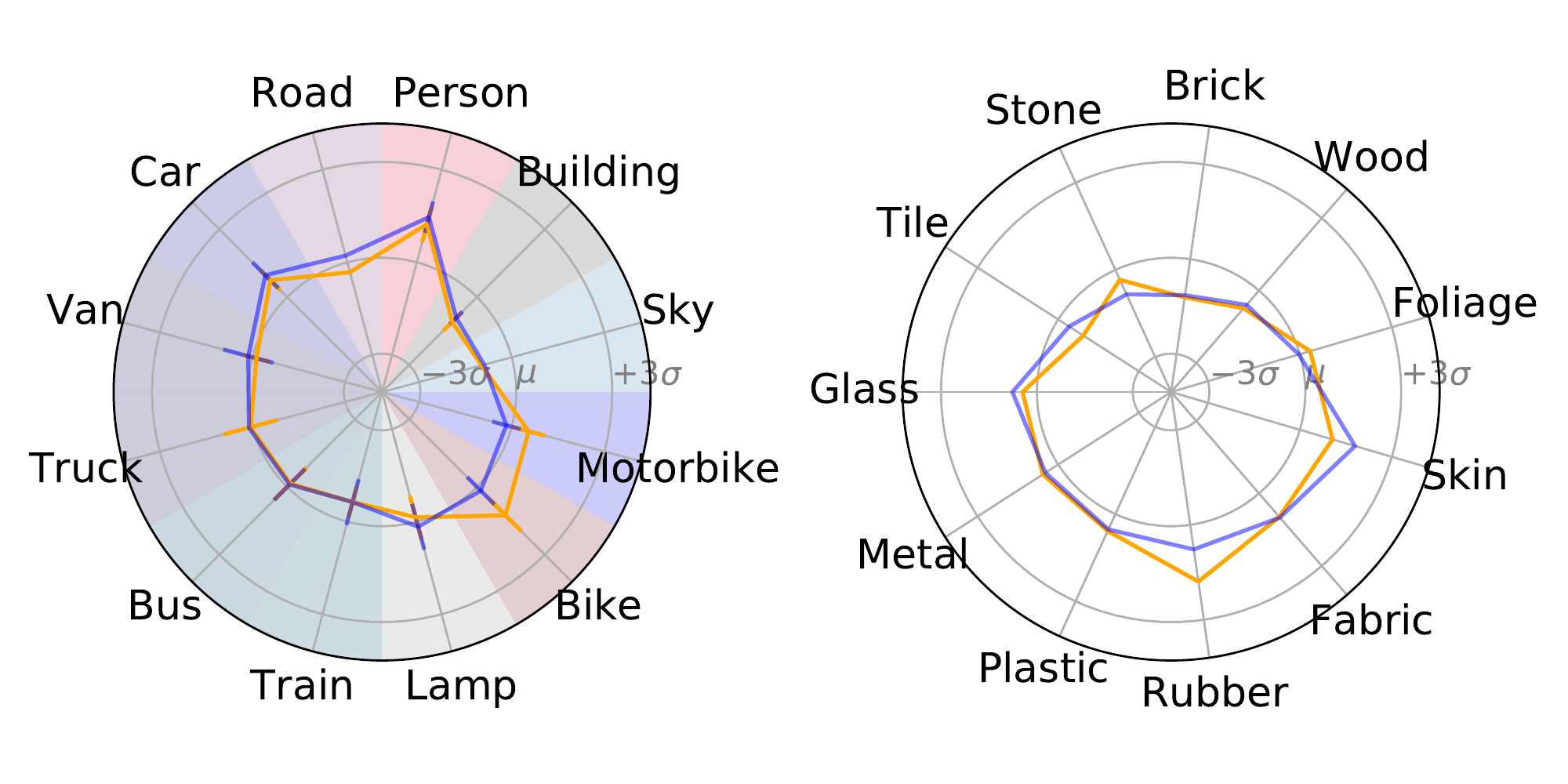}
        \end{center}
    \end{subfigure}
    
    \begin{subfigure}[t]{0.4\linewidth}
        \begin{center}
           \includegraphics[width=\linewidth]{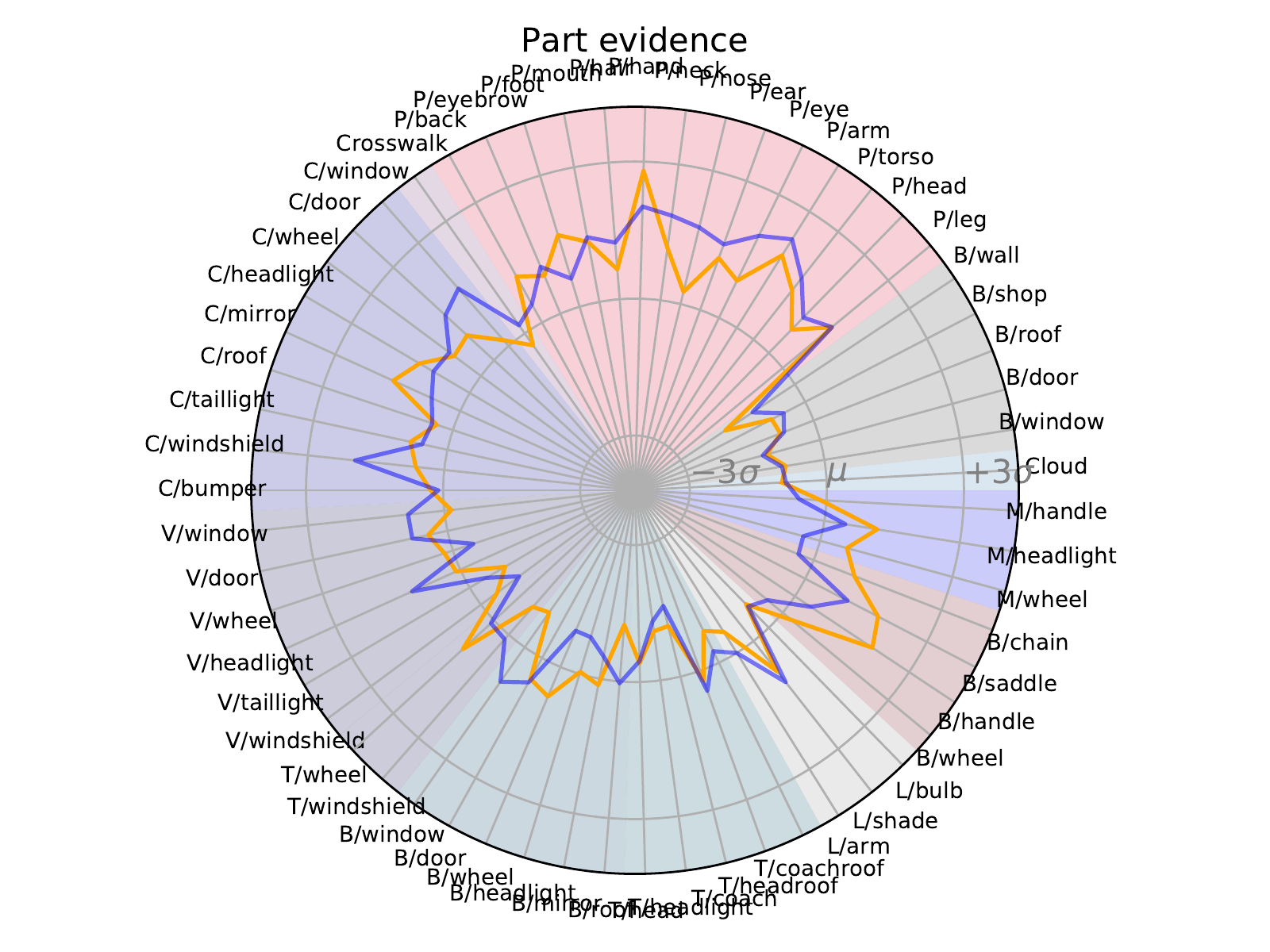}
        \end{center}
        \vspace{-15pt}
        \caption{Individual part activations}
    \end{subfigure}
    
    \begin{subfigure}[t]{0.59\linewidth}
        \begin{center}
           \includegraphics[width=\linewidth]{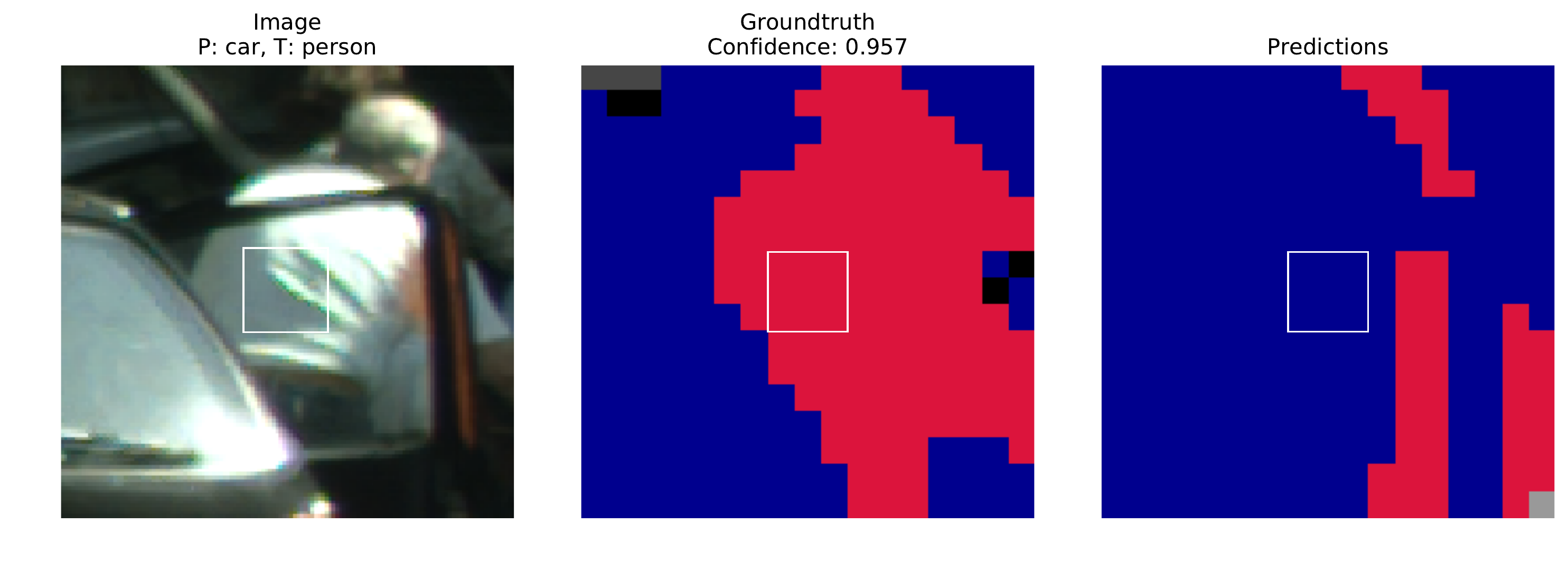}
        \end{center}
    \end{subfigure}
    \begin{subfigure}[t]{0.4\linewidth}
        \begin{center}
           \includegraphics[width=\linewidth]{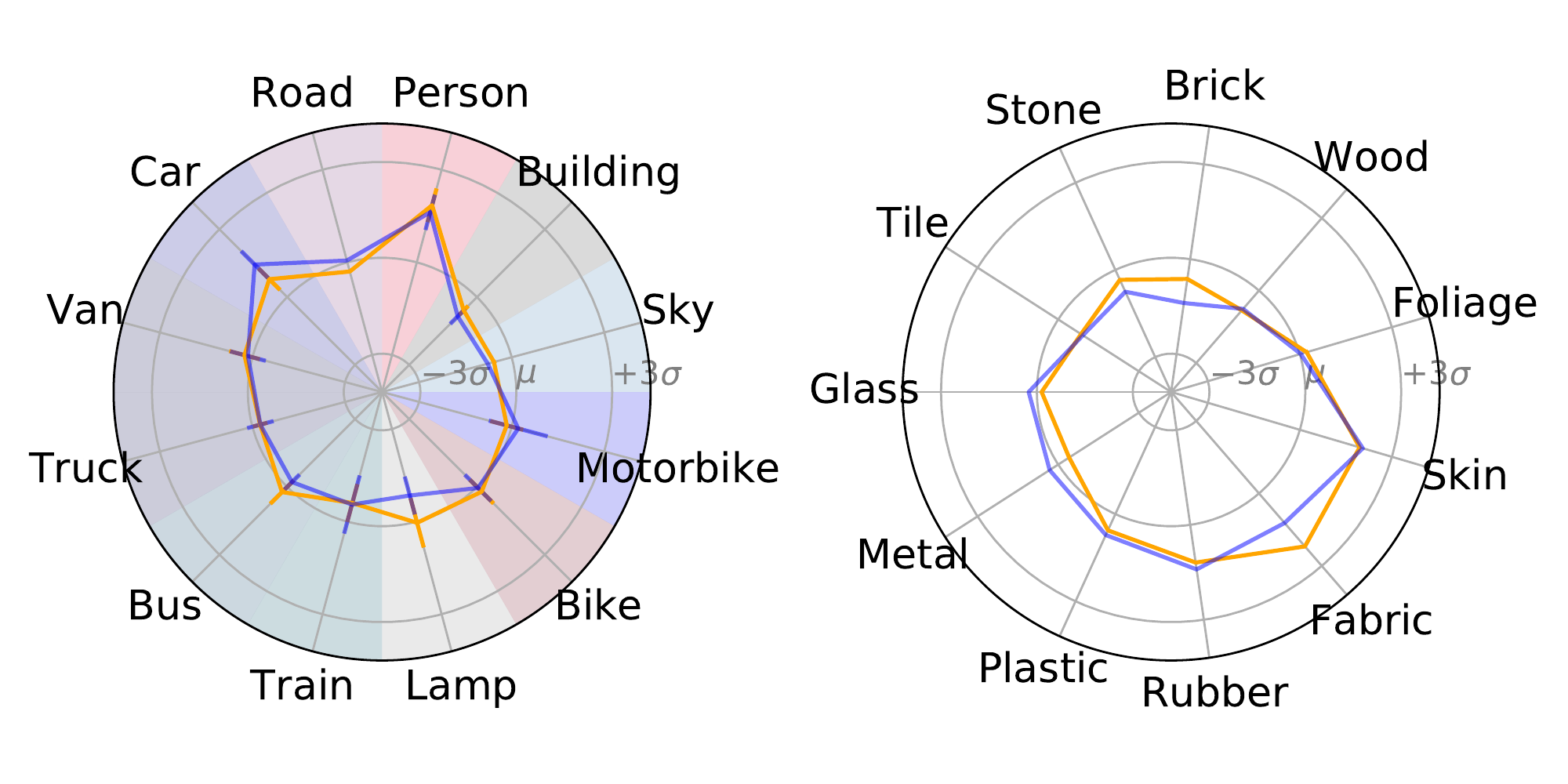}
        \end{center}
    \end{subfigure}
    
    \begin{subfigure}[t]{0.4\linewidth}
        \begin{center}
           \includegraphics[width=\linewidth]{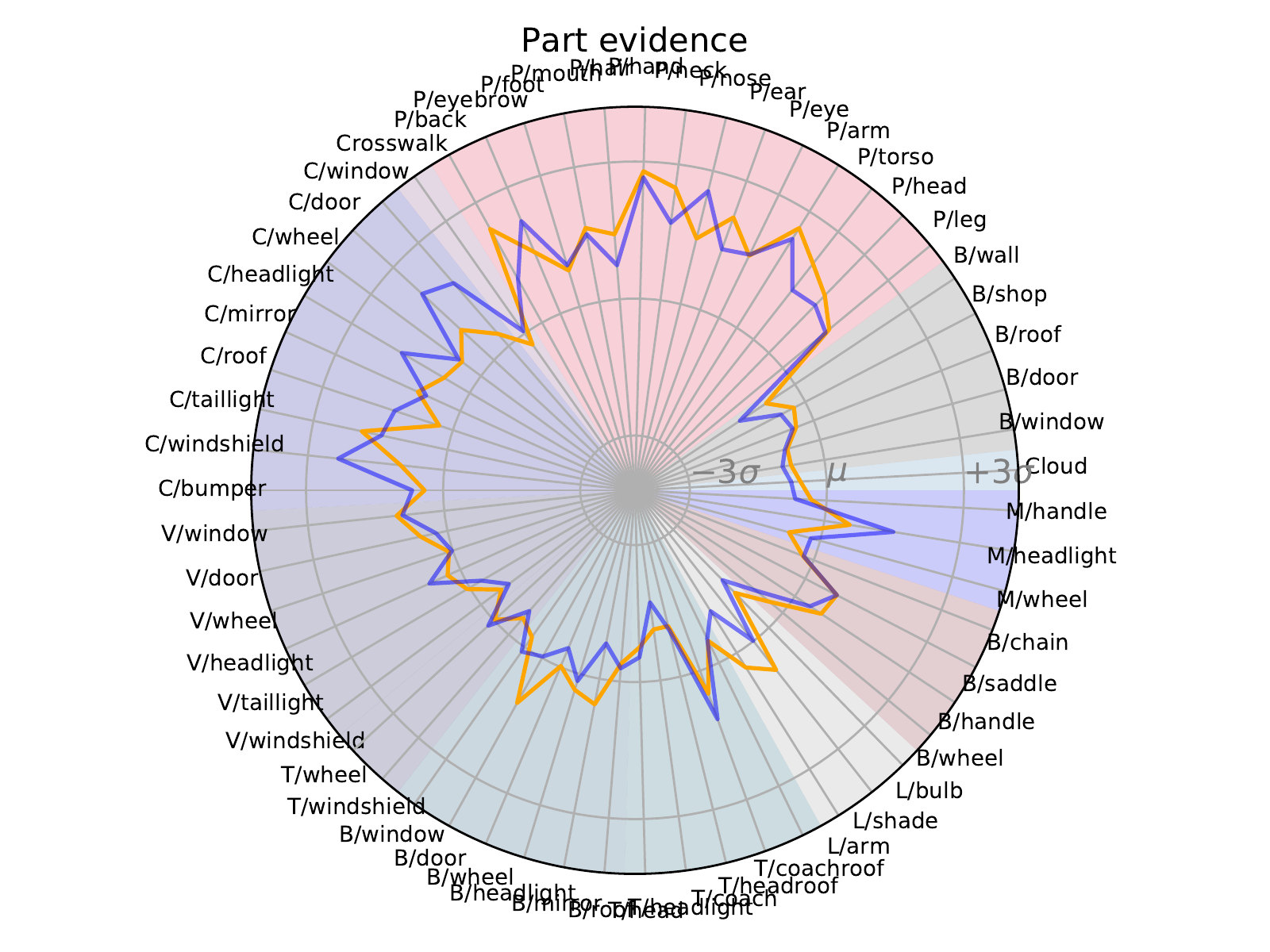}
        \end{center}
        \vspace{-15pt}
        \caption{Individual part activations}
    \end{subfigure}
    \caption{Selection of $3$ samples from a cluster of size $101$ related to error pixels in $9$ different images.}
    \label{fig:ex_person_as_car}
\end{figure}

\begin{figure}
    \centering
    \begin{subfigure}[t]{0.59\linewidth}
        \begin{center}
           \includegraphics[width=\linewidth]{figures/error_cluster/2135_fns_building_as_person/0007_img_gt_pred.pdf}
        \end{center}
    \end{subfigure}
    \begin{subfigure}[t]{0.4\linewidth}
        \begin{center}
           \includegraphics[width=\linewidth]{figures/error_cluster/2135_fns_building_as_person/0007_part_and_mat.pdf}
        \end{center}
    \end{subfigure}
    
    \begin{subfigure}[t]{0.4\linewidth}
        \begin{center}
           \includegraphics[width=\linewidth]{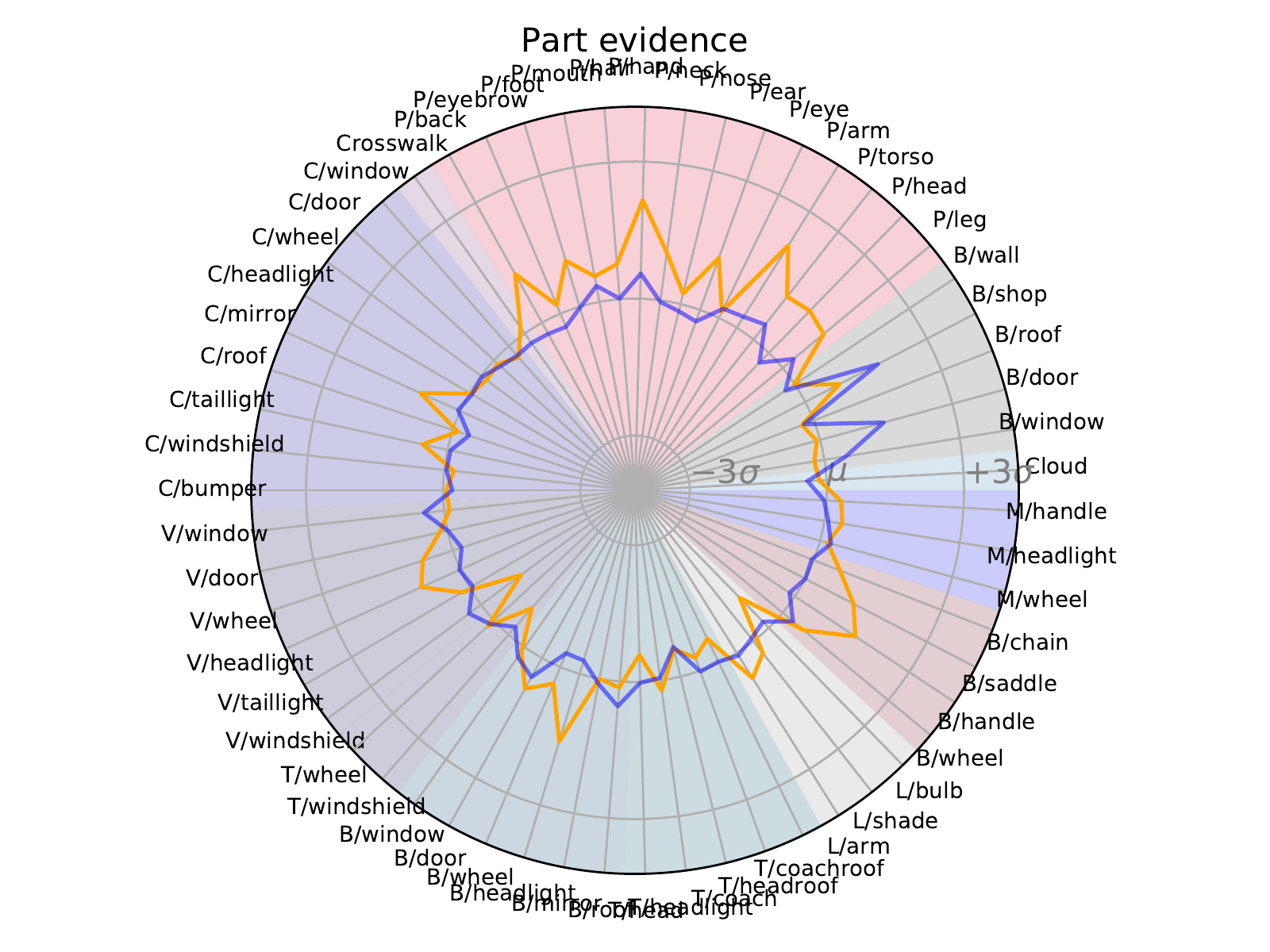}
        \end{center}
        \vspace{-15pt}
        \caption{Individual part activations}
    \end{subfigure}
    
    \begin{subfigure}[t]{0.59\linewidth}
        \begin{center}
           \includegraphics[width=\linewidth]{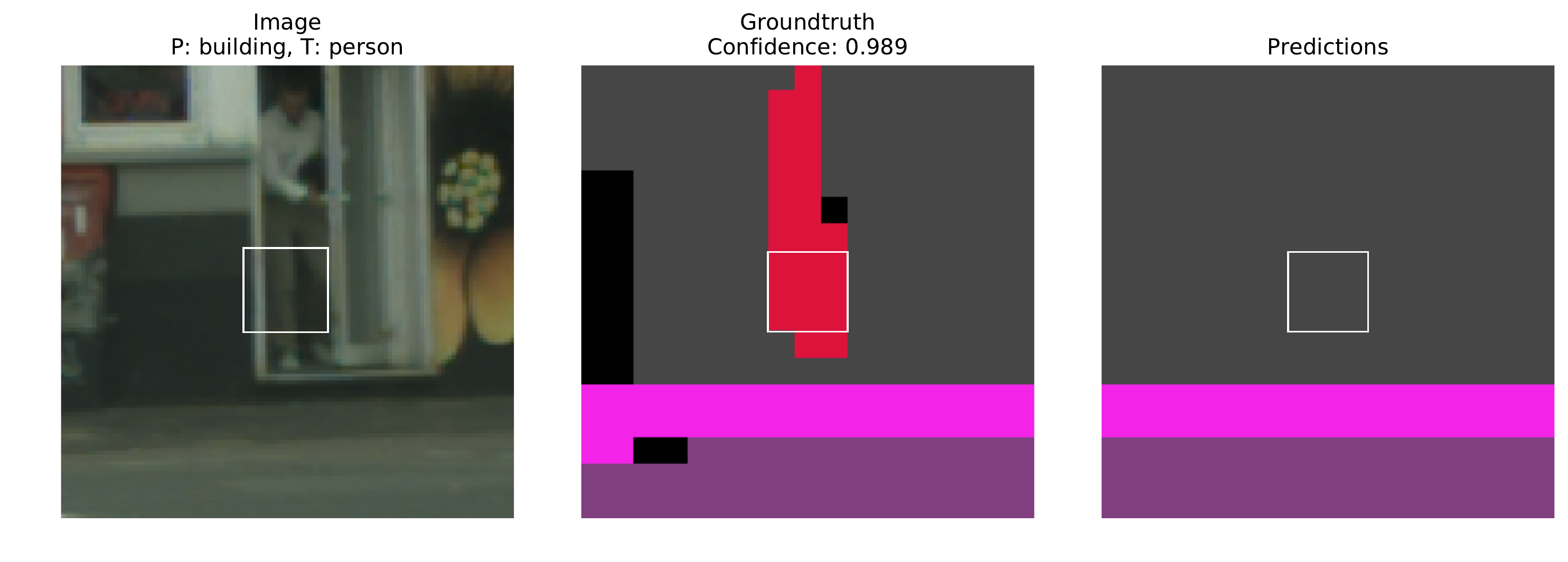}
        \end{center}
    \end{subfigure}
    \begin{subfigure}[t]{0.4\linewidth}
        \begin{center}
           \includegraphics[width=\linewidth]{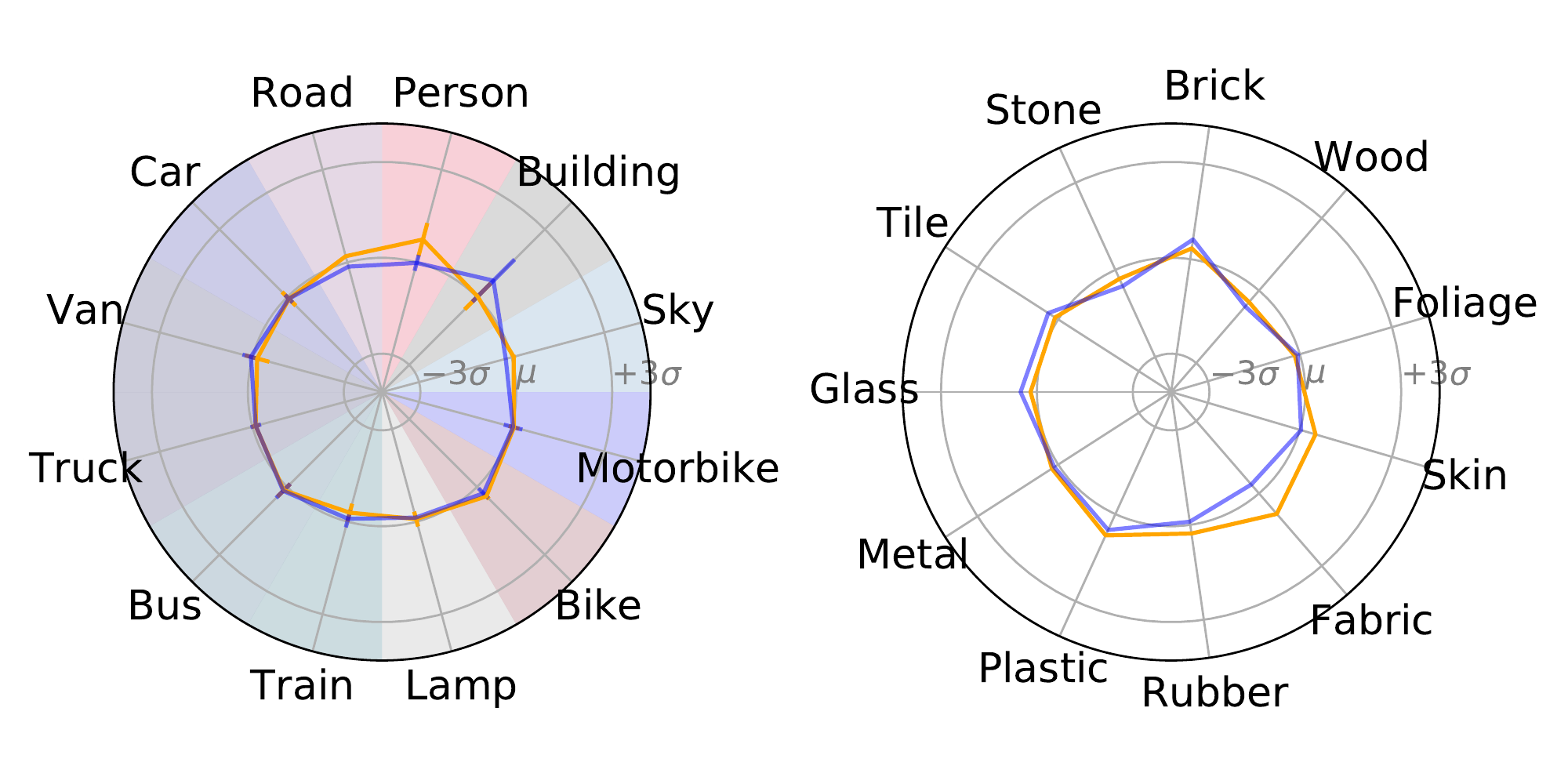}
        \end{center}
    \end{subfigure}
    
    \begin{subfigure}[t]{0.4\linewidth}
        \begin{center}
           \includegraphics[width=\linewidth]{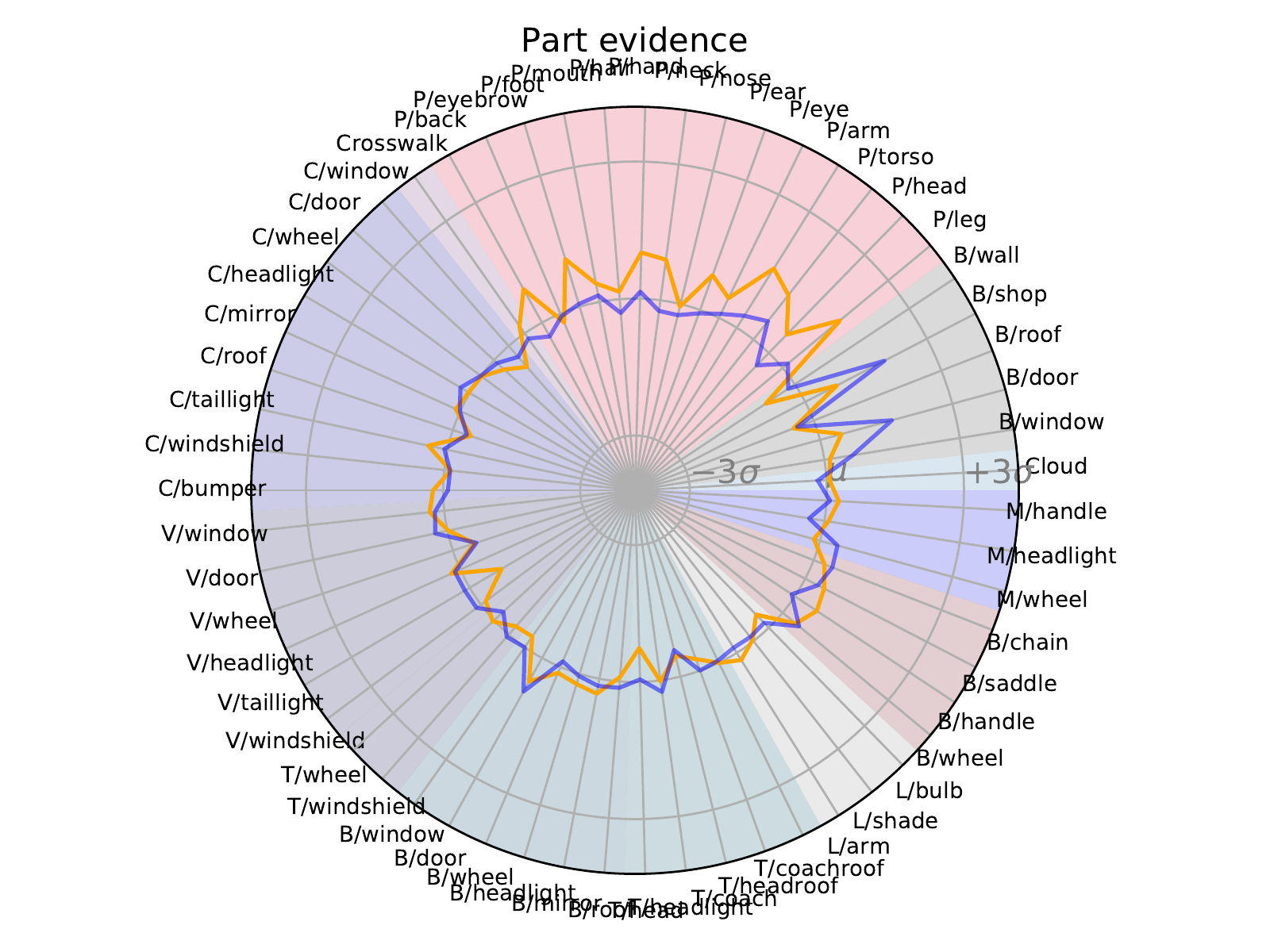}
        \end{center}
        \vspace{-15pt}
        \caption{Individual part activations}
    \end{subfigure}
    
    \begin{subfigure}[t]{0.59\linewidth}
        \begin{center}
           \includegraphics[width=\linewidth]{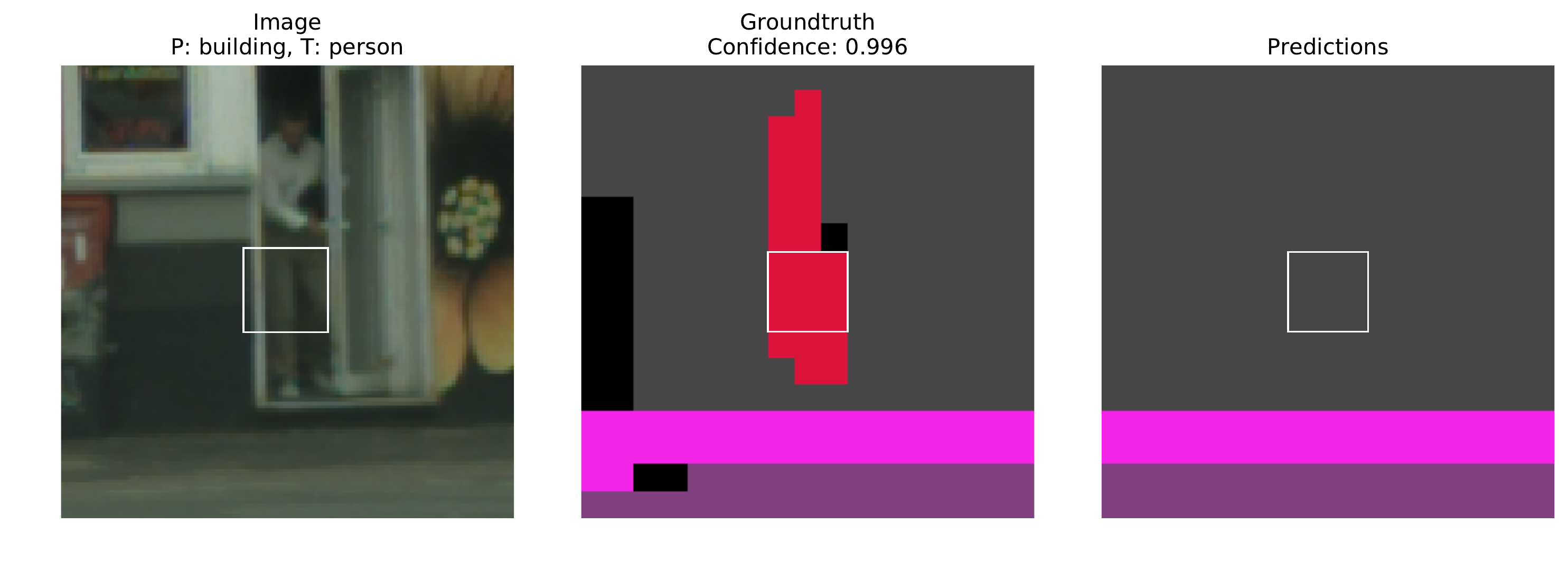}
        \end{center}
    \end{subfigure}
    \begin{subfigure}[t]{0.4\linewidth}
        \begin{center}
           \includegraphics[width=\linewidth]{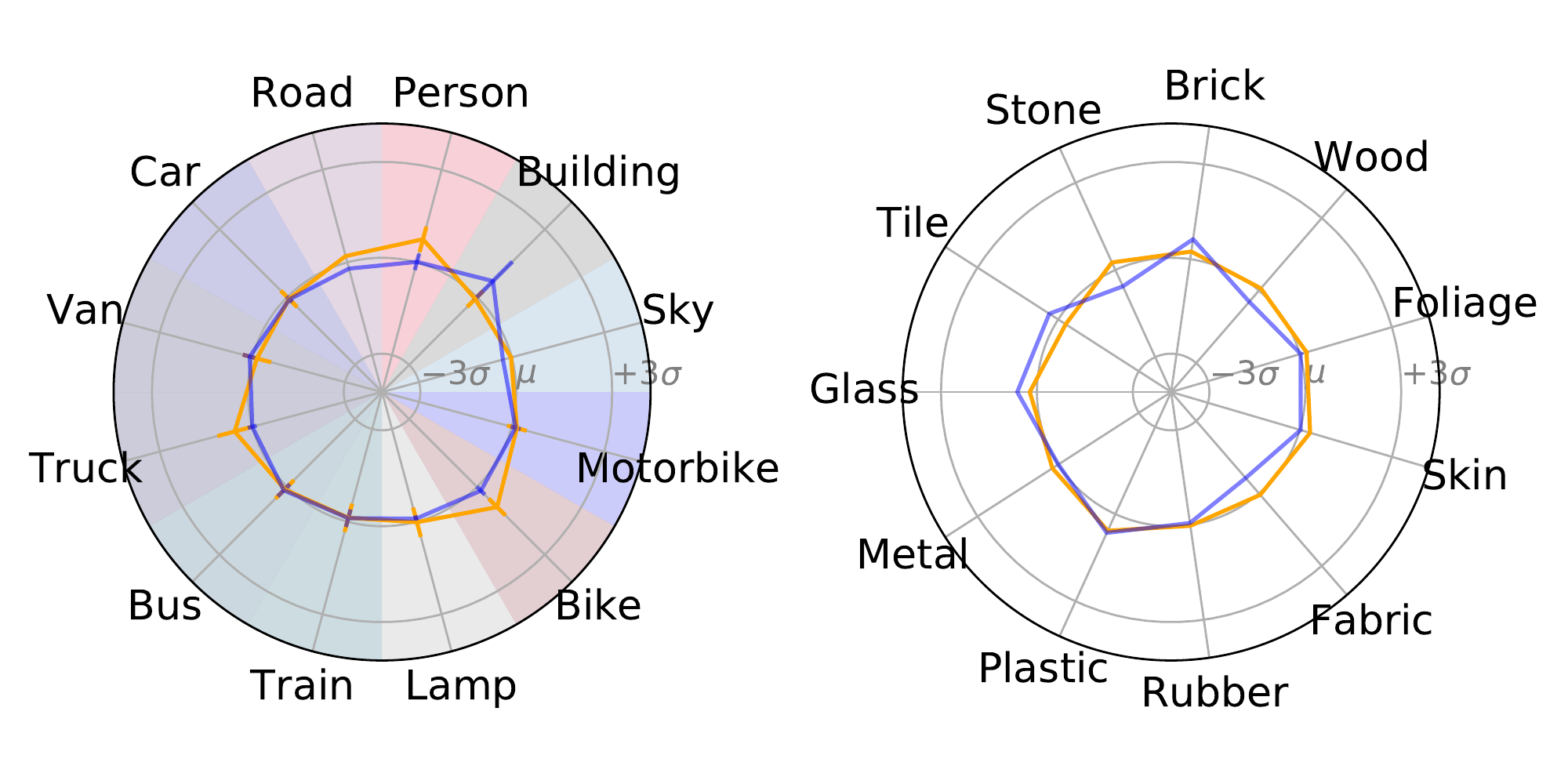}
        \end{center}
    \end{subfigure}
    
    \begin{subfigure}[t]{0.4\linewidth}
        \begin{center}
           \includegraphics[width=\linewidth]{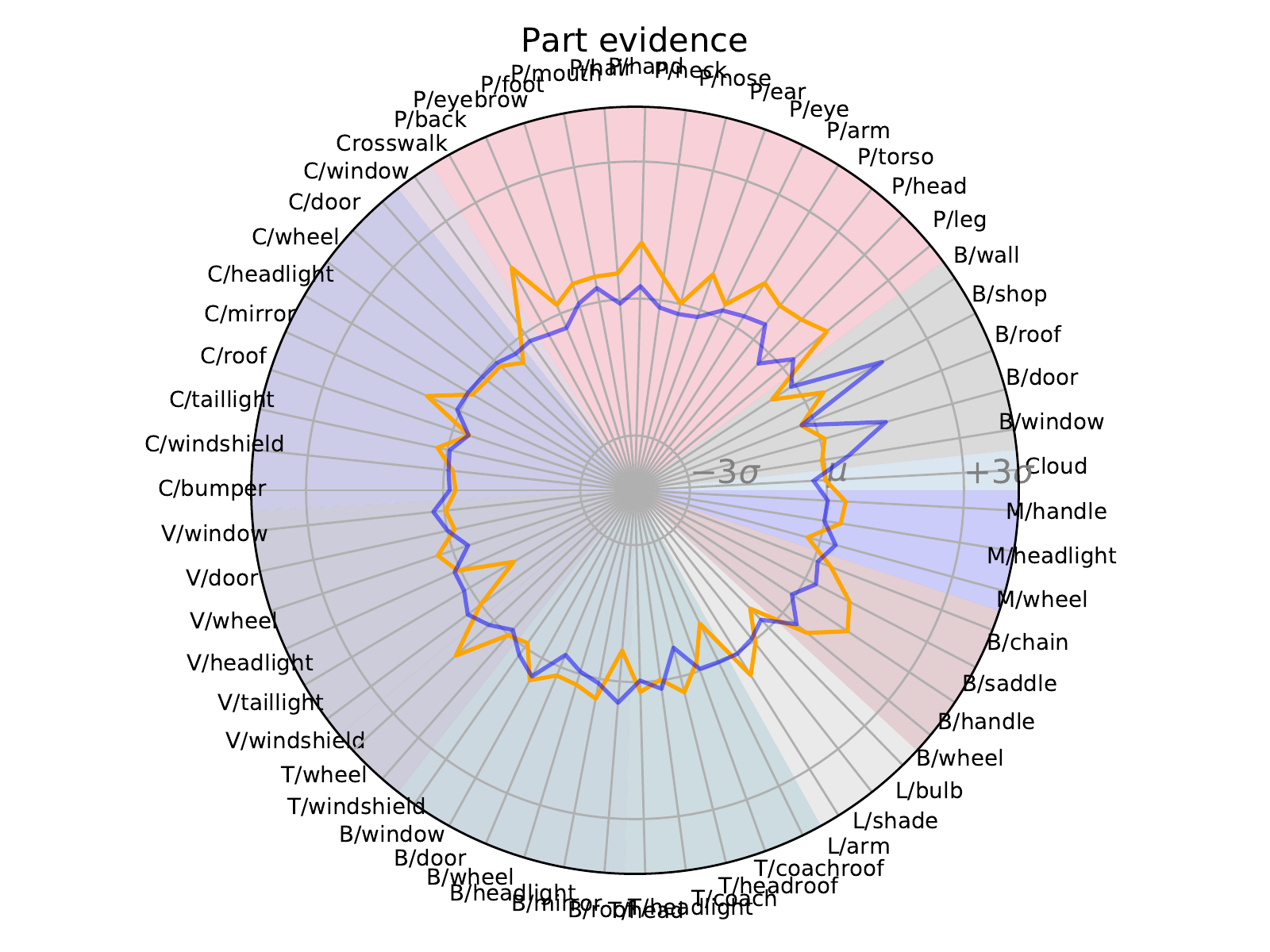}
        \end{center}
        \vspace{-15pt}
        \caption{Individual part activations}
    \end{subfigure}
    \caption{Selection of $3$ samples from a cluster of size $12$ related to error pixels in a single image.}
    \label{fig:ex_building_as_person}
\end{figure}

\begin{figure}
    \centering
    \begin{subfigure}[t]{0.59\linewidth}
        \begin{center}
           \includegraphics[width=\linewidth]{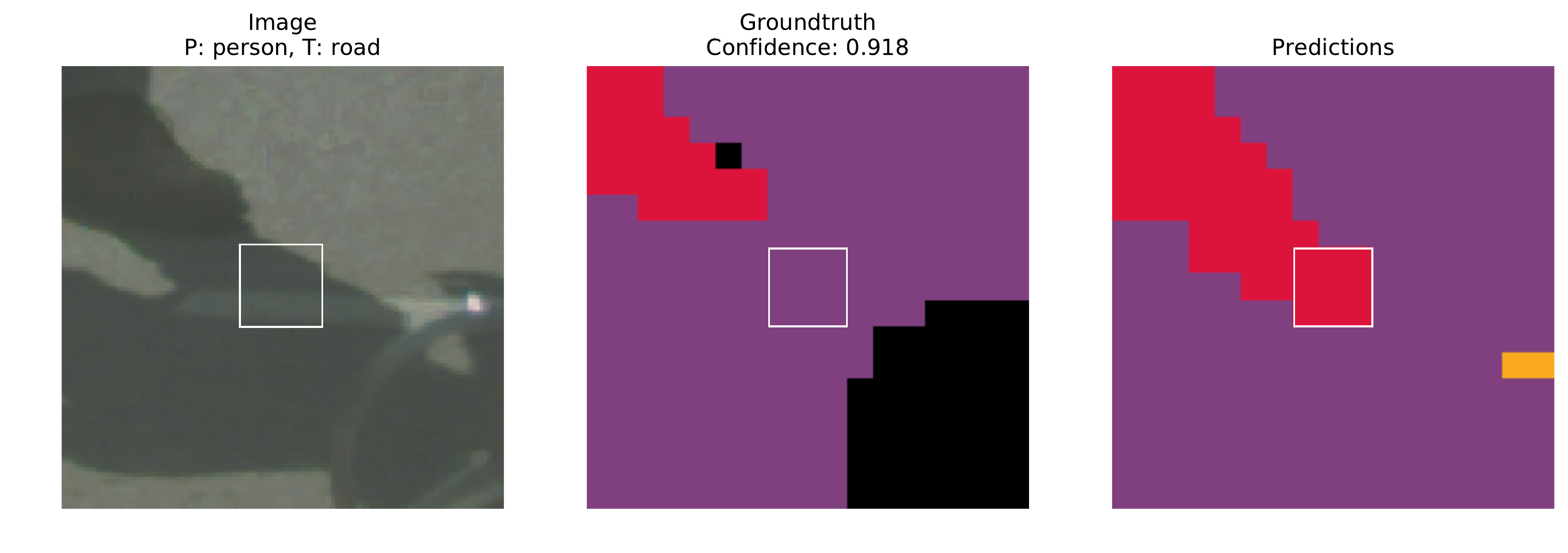}
        \end{center}
    \end{subfigure}
    \begin{subfigure}[t]{0.4\linewidth}
        \begin{center}
           \includegraphics[width=\linewidth]{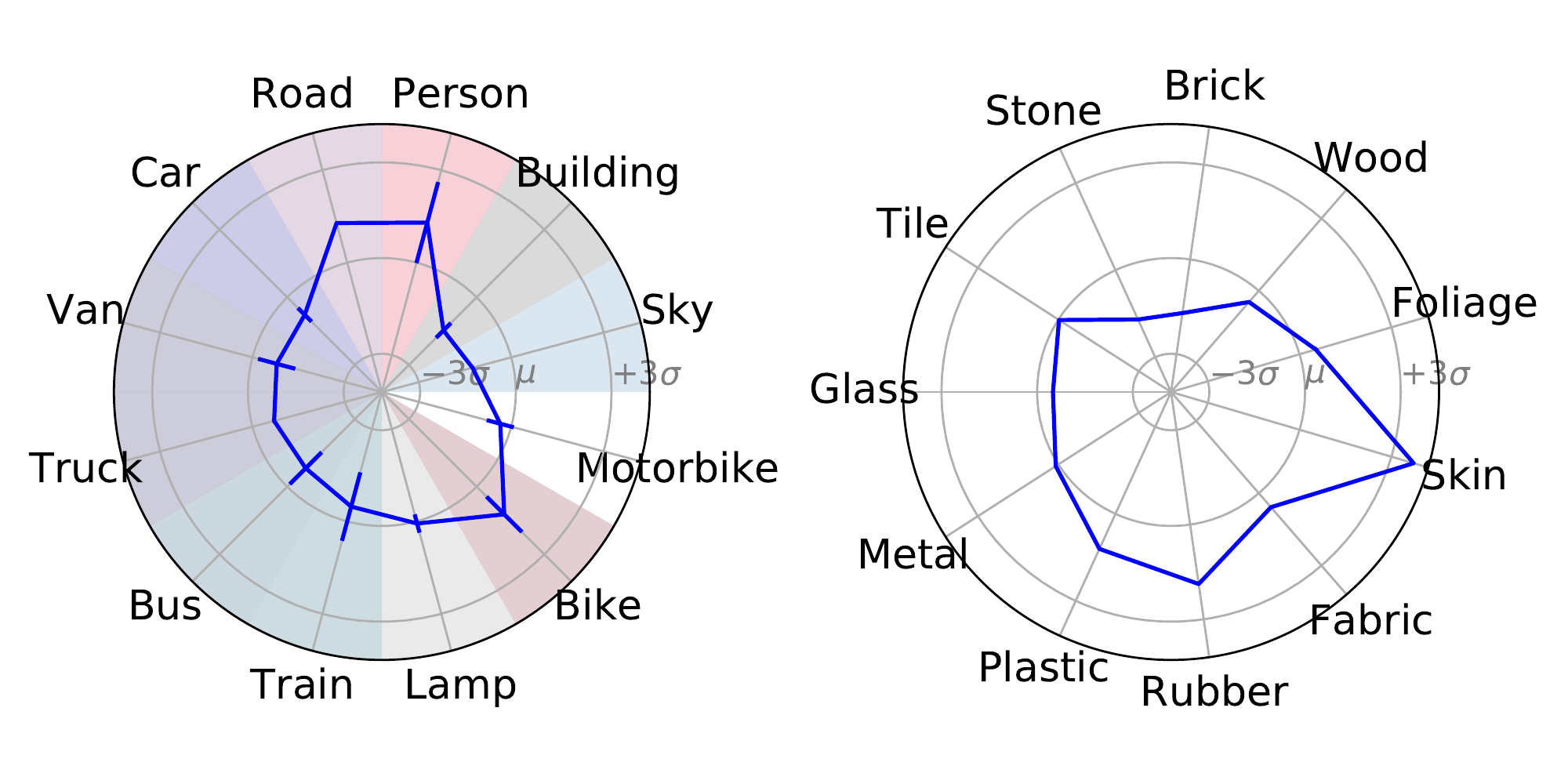}
        \end{center}
    \end{subfigure}
    
    \begin{subfigure}[t]{0.4\linewidth}
        \begin{center}
           \includegraphics[width=\linewidth]{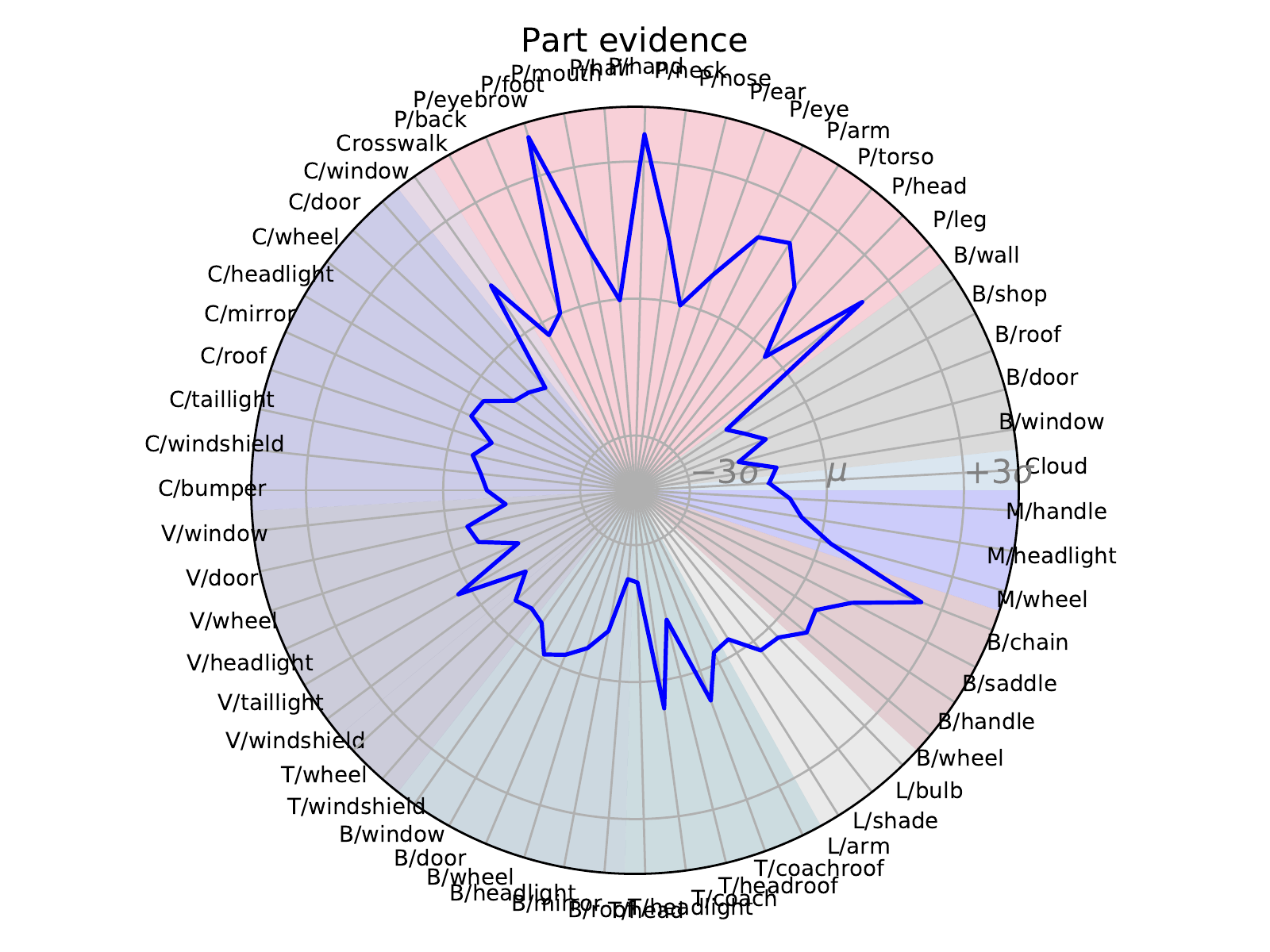}
        \end{center}
        \vspace{-15pt}
        \caption{Individual part activations}
    \end{subfigure}
    
    \begin{subfigure}[t]{0.59\linewidth}
        \begin{center}
           \includegraphics[width=\linewidth]{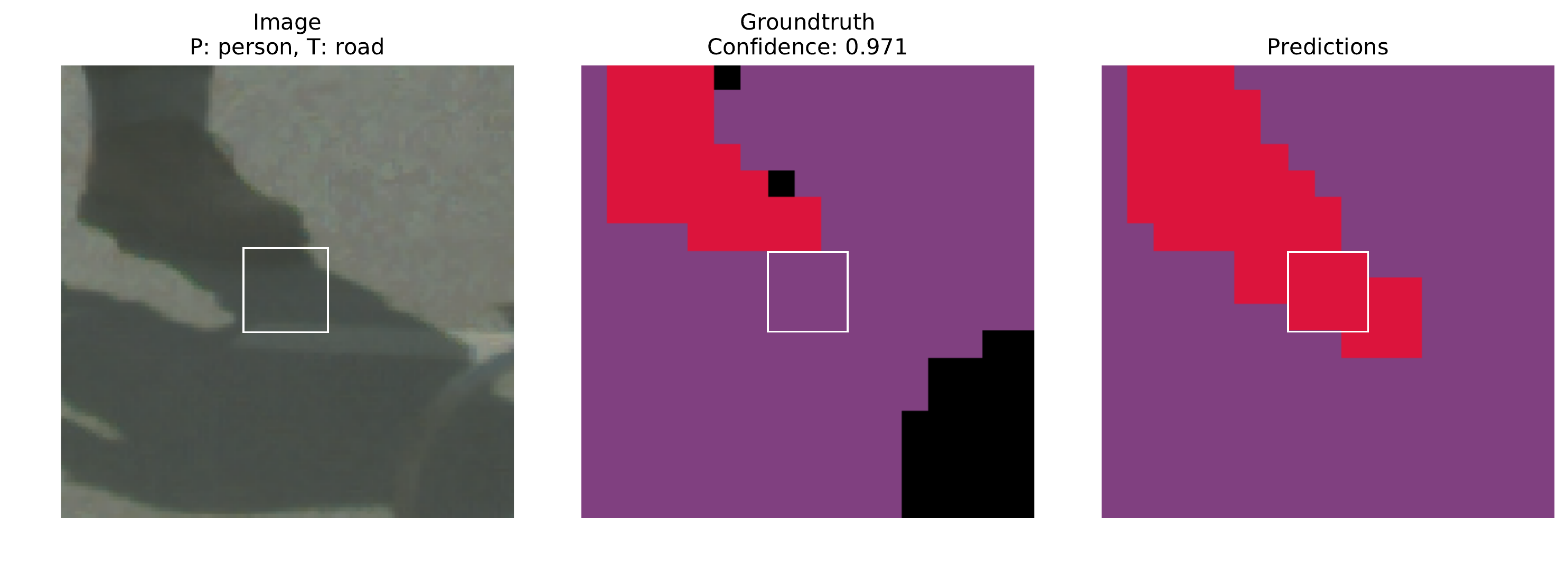}
        \end{center}
    \end{subfigure}
    \begin{subfigure}[t]{0.4\linewidth}
        \begin{center}
           \includegraphics[width=\linewidth]{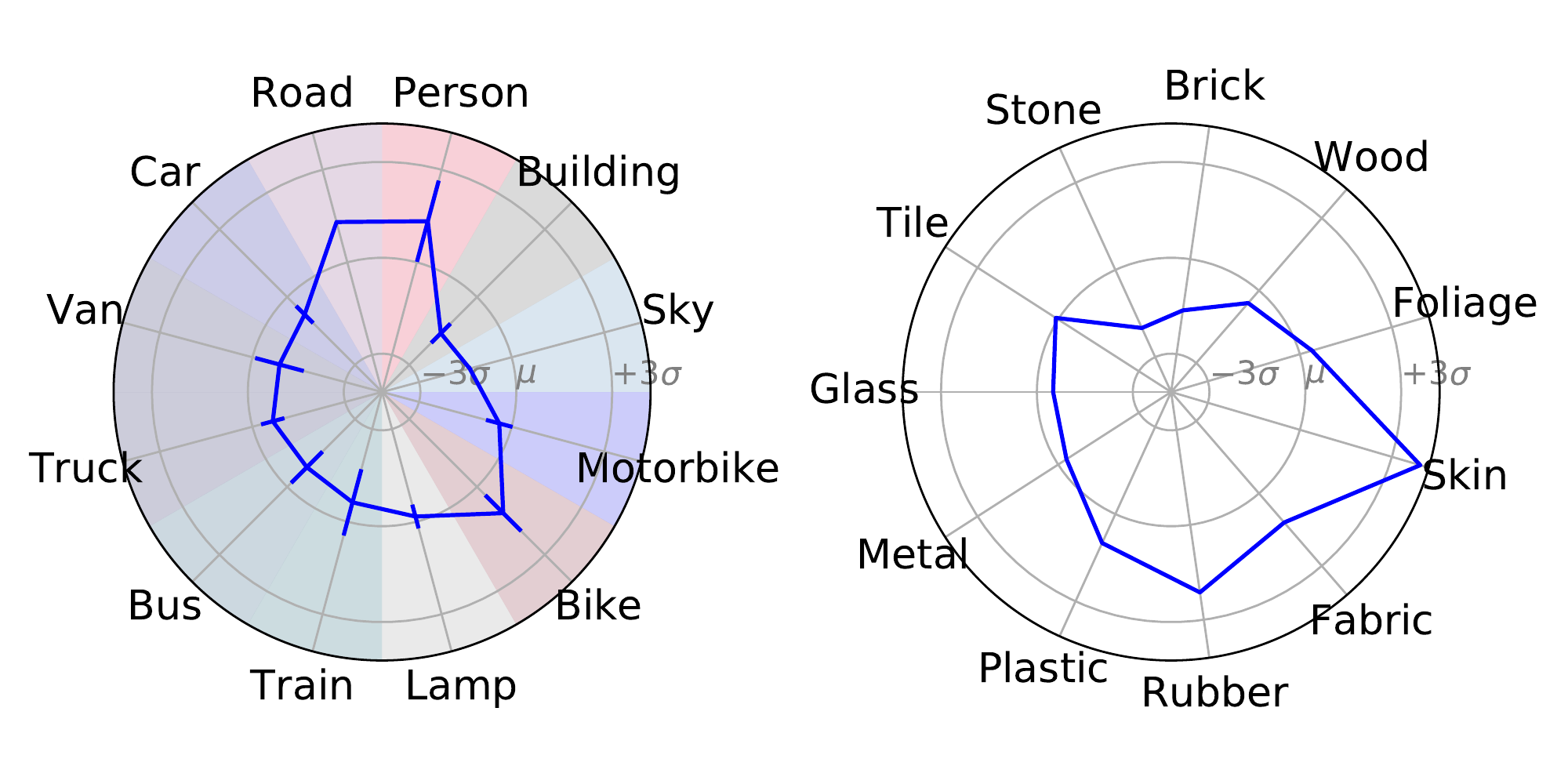}
        \end{center}
    \end{subfigure}
    
    \begin{subfigure}[t]{0.4\linewidth}
        \begin{center}
           \includegraphics[width=\linewidth]{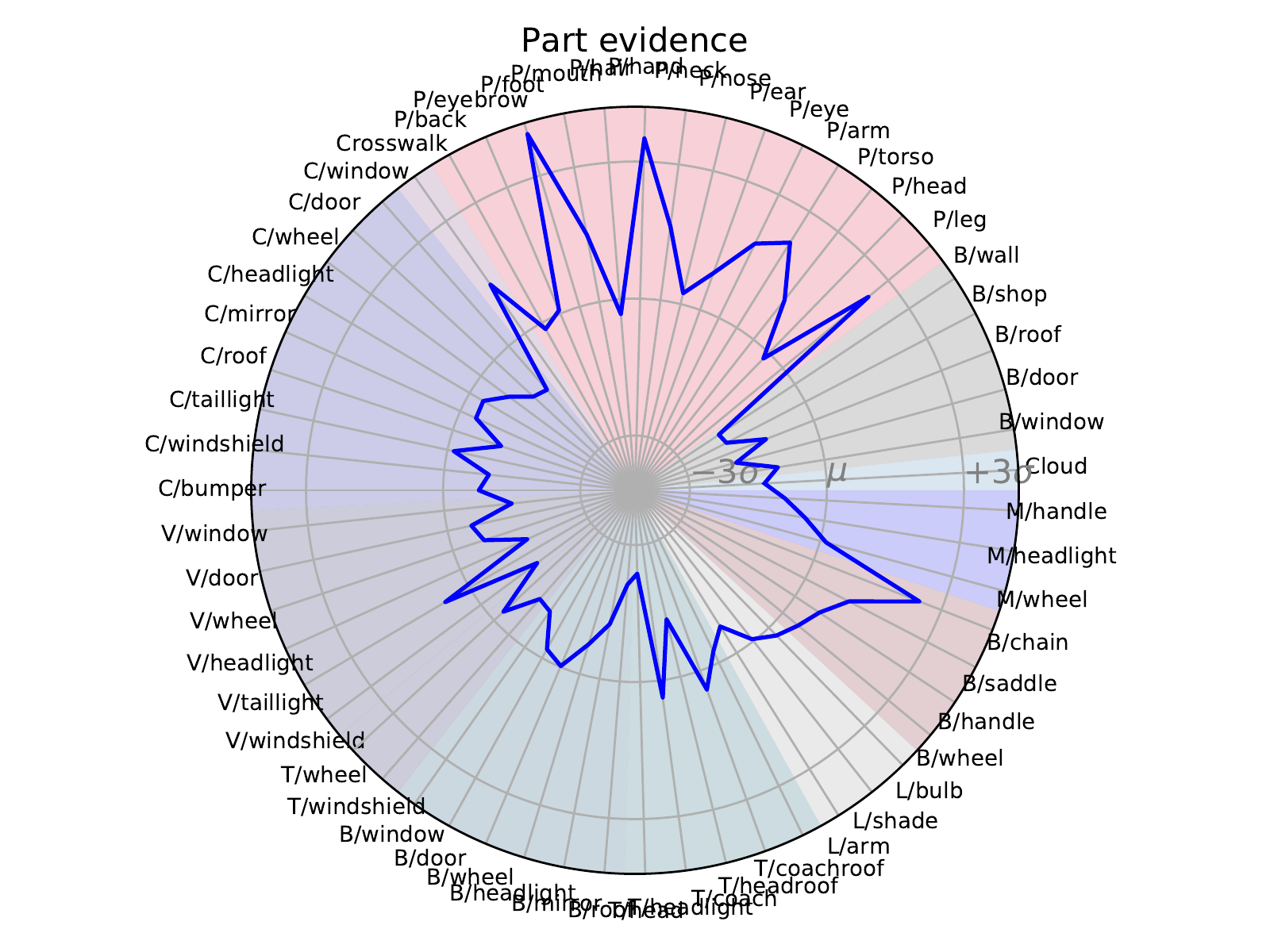}
        \end{center}
        \vspace{-15pt}
        \caption{Individual part activations}
    \end{subfigure}

    \caption{Selection of $2$ samples from a cluster of size $15$ related to error pixels in a single image. We did not determine the counterfactual for this example.}
    \label{fig:ex_road_as_person}
\end{figure}

As supplement to the visualization of Semantic Bottleneck (SB) activations in figure~\ref{fig:errors} in the paper, we list the following plots in no particular order to add further examples and especially to show the individual concept part activations.
\begin{itemize}
    \item \textbf{Figure~\ref{fig:ex_sign_as_person}:} Construction site beacons misclassified due to overhelming evidence for material \textit{skin} and person related concepts.
    In particular see the individual part concepts \textit{foot} and \textit{leg} at $11$ o'clock and $2$ o'clock on the radar plots.
    
    \item \textbf{Figure~\ref{fig:ex_person_as_car}:} Person standing between cars misclassified as car, as the average \textit{car} activation is higher than the average \textit{person} activation.
    Also observe that the orange line indicating the counterfactual shows that the SBN is classifying the evidence correctly if the amplitude of activations is switched.
    
    \item \textbf{Figure~\ref{fig:ex_building_as_person}:} Person standing in a door misclassified as building. It is particularly evident that the SB carries no to very little evidence for any person concept. Especially surprising is the lack of evidence for the concept \textit{leg}.
    
    \item \textbf{Figure~\ref{fig:ex_road_as_person}:} A shadow of a person standing on the road is misclassified as person. The SB shows clear evidence for person concepts, especially \textit{foot} ($11$ o'clock on the radar plot).
\end{itemize}

\end{document}